\documentclass[twoside,11pt]{article}

\usepackage{blindtext}

\usepackage{jmlr2e}

\usepackage[utf8]{inputenc}
\usepackage{enumitem}
\usepackage{subcaption} 

\usepackage{makecell}

\usepackage{amsmath} 
\usepackage{amsfonts} 
\usepackage{mathtools} 
\usepackage{bm}
\usepackage[thinc]{esdiff}

\usepackage{algorithm}
\usepackage{algpseudocode} 
\usepackage{xcolor}

\definecolor{mygreen}{RGB}{0,128,0}
\definecolor{cobalt}{rgb}{0.0, 0.28, 0.67}

\usepackage{hyperref} 
\usepackage{tikz}

\newcommand{\PS}{\operatorname{\mathcal{PS} }}

\usepackage[textwidth=1.8cm, textsize=scriptsize]{todonotes}

\firstpageno{1}

\begin{document}

\title{Generalized Bayesian deep reinforcement learning}

\author{\name Shreya Sinha Roy \email shreya.sinha-roy@warwick.ac.uk\\
\name Richard G. Everitt \email richard.everitt@warwick.ac.uk\\
\name Christian P. Robert\thanks{Also affiliated with CEREMADE, Université Paris Dauphine PSL, France} \email c.a.m.robert@warwick.ac.uk\\
\name Ritabrata Dutta \email ritabrata.dutta@warwick.ac.uk\\
\addr Department of Statistics, University of Warwick\\
Coventry, CV4 7AL\\
United Kingdom}

\editor{} 

\maketitle

\begin{abstract}
Bayesian reinforcement learning (BRL) is a method that merges principles from Bayesian statistics and reinforcement learning to make optimal decisions in uncertain environments. As a model-based RL method, it has two key components: (1) inferring the posterior distribution of the model for the data-generating process (DGP) and (2) policy learning using the learned posterior. We propose to model the dynamics of the unknown environment through deep generative models, assuming Markov dependence. In the absence of likelihood functions for these models, we train them by learning a generalized predictive-sequential (or prequential) scoring rule (SR) posterior. We used sequential Monte Carlo (SMC) samplers to draw samples from this generalized Bayesian posterior distribution. In conjunction, to achieve scalability in the high-dimensional parameter space of the neural networks, we use the gradient-based Markov kernels within SMC. To justify the use of the prequential scoring rule posterior, we prove a Bernstein-von Mises-type theorem. 
For policy learning, we propose expected Thompson sampling (ETS) to learn the optimal policy by maximising the expected value function with respect to the posterior distribution. This improves upon traditional Thompson sampling (TS) and its extensions, which utilize only one sample drawn from the posterior distribution. This improvement is studied both theoretically and using simulation studies, assuming a discrete action space. 
Finally, we successfully extended our setup for a challenging problem with a continuous action space without theoretical guarantees.
\end{abstract}


\section{Introduction}
\label{sec:introduction}

Effective learning and decision-making within perpetually changing dynamic systems are crucial for applications like controlling automated machinery and enabling robotic navigation. Reinforcement learning (RL) stands out as a potent tool in these domains, allowing the agents to acquire knowledge through trial and error and responses from the environment. Its versatility has resulted in its widespread utilization in a variety of sectors, including automated vehicles \citep{guan2020centralized}, robotics \citep{kormushev2013reinforcement}, healthcare \citep{yu2021reinforcement}, finance \citep{deng2016deep}, various applications of natural language processing (NLP) \citep{uc2023survey}, recommendation systems \citep{chen2023deep}, and so on.

Under the classical framework, an
RL task can be expressed using a Markov decision process (MDP) \citep{sutton2018reinforcement}. 
The multitude of RL algorithms existing in the literature can be broadly segregated into two categories: those reliant on a model (`model-based' algorithms) and those that facilitate policy learning in a `model-free' manner. Model-free algorithms like \textit{Q-learning algorithms} \citep{clifton2020q} and
\textit{policy gradient methods} \citep{sutton2018reinforcement} learn directly from the history generated by real-time interactions. In contrast, model-based approaches such as dynamic programming \citep{sutton2018reinforcement},
Monte Carlo tree search \citep{coulom2006efficient},
PILCO \citep{deisenroth2011pilco} etc. use knowledge about the environment to design a policy.

When a reliable model of the environment is available or can be learned, model-based RL approaches \citep{moerland2023model} tend to be significantly more sample efficient than model-free methods. In this paradigm, Thompson sampling (TS) \citep{russo2018tutorial}, also known as posterior sampling for RL (PSRL) \citep{strens2000bayesian}, offers an effective solution to the exploration-exploitation trade-off and enjoys strong theoretical guarantees, including provable regret bounds \citep{osband2013more, ouyang2017learning}. However, its application has been largely restricted to settings with simple MDPs with a tractable likelihood function of the parameters of the model, allowing posterior inference over model parameters.

Recent advances in model-based RL have leveraged powerful neural network-based predictive models \citep{nagabandi2018neural, kaiser2019model} and conditional GANs \citep{charlesworth2020plangan, zhao2021model} to capture complex environment dynamics. Although highly expressive, these models often lack a tractable likelihood function, making training and inference challenging. Previous work addresses this by approximating divergences (e.g., through adversarial training \citep{goodfellow2020generative}) or using surrogate scores such as Fisher information \citep{gurney2018introduction}. However, the absence of a well-defined likelihood hinders posterior inference, limiting the applicability of Thompson sampling and its Bayesian extensions.

Our main aim in this work is \textit{to provide a tool to perform Thompson sampling for model-based RL when the models considered are deep generative models}. Traditional likelihood-free methods such as approximate Bayesian computation (ABC) have been used for TS in this context \citep{dimitrakakis2013abc}, but their scalability is limited due to the curse of dimensionality. An alternative is to use scoring rules (SR)\citep{gneiting2007strictly}, which allow inference when the model lacks a tractable likelihood but is easy to simulate from. Recently, predictive-sequential or prequential scoring rules \citep{dawid1984present, dawid1999prequential} have shown promise in the training of deep generative models for forecasting tasks \citep{pacchiardi2024probabilistic}. Building on this, we construct a generalized posterior \citep{bissiri2016general} using the prequential score as a surrogate for the log-likelihood of MDPs modeled by generative networks. TS can then be performed using this generalized posterior. In particular, we employ sequential Monte Carlo (SMC) \citep{del2006sequential} samplers enhanced with preconditioned gradient-based Markov kernels \cite{chen2016bridging} to efficiently explore and sample from the posterior distributions.

Thompson sampling (TS) is a simple yet effective policy learning strategy that relies on a single sample from the posterior over model parameters. Although some recent work \citep{dimitrakakis2022decision} suggests that the use of multiple samples can improve performance, this is still underexplored. Motivated by this, we propose expected Thompson sampling (ETS), a policy learning approach that optimizes the expected action-value function in the posterior distribution. In practice, this expectation is approximated using multiple posterior samples. Through both theoretical analysis and simulations, we demonstrate that the regret of ETS-based policies decreases as the number of samples increases.

Therefore, our proposed approach combines the robustness of scoring rules for Bayesian inference, employs SMC for high dimensional parameter spaces, and provides a strategy to design a policy, by potentially better handling model uncertainty. The main contributions of our work are as follows:
\begin{itemize}
\item A scalable, likelihood-free TS framework for deep generative models in RL.
\item A theoretical bound on the value function approximation error under ETS.
\item An efficient SMC-based implementation of ETS that scales to high-dimensional settings.
 
\end{itemize}

\begin{figure}
    \centering
    \includegraphics[width = 0.8\linewidth]{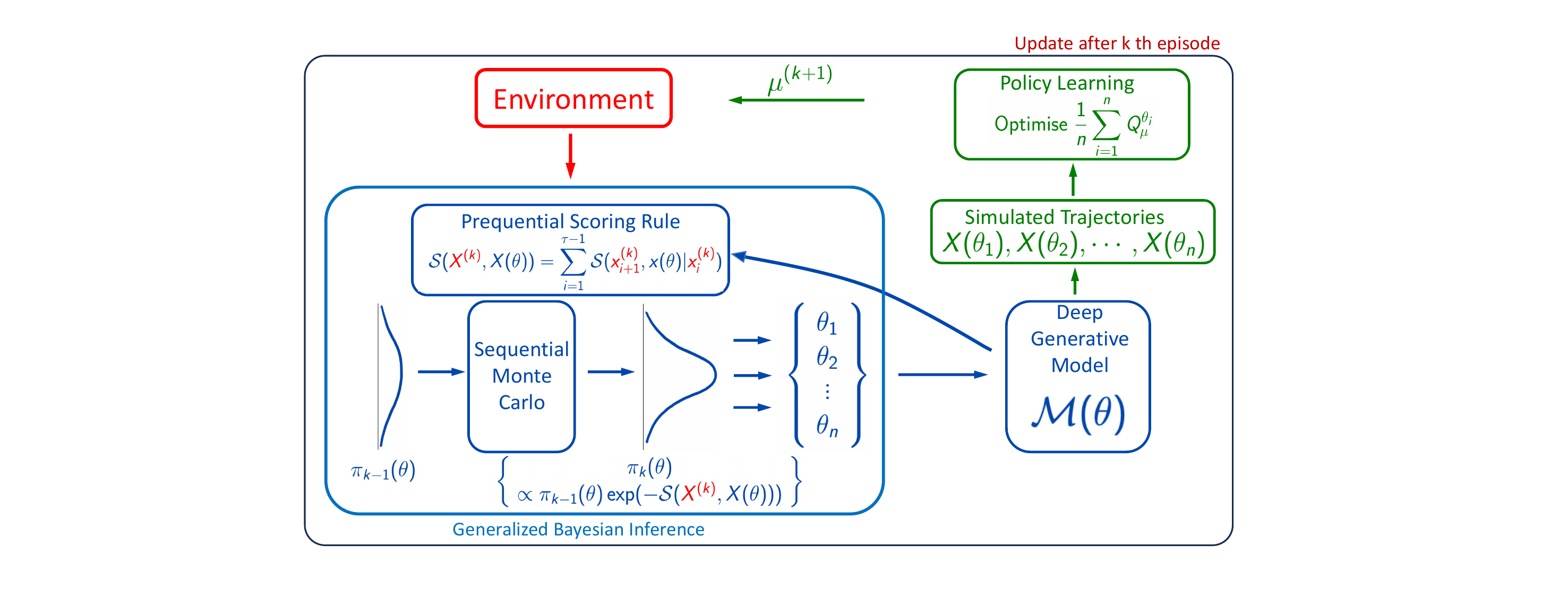}
    \caption{\textbf{Generalized Bayesian deep RL}: The diagram illustrates the episodic posterior and policy update process for the $k$th episode ($k = 1, 2, \ldots$) with episode length $\tau$. Starting with a prior $\pi_0(\theta)$ on model parameters, the generalized prequential posterior $\pi_k(\theta) \propto \pi_{k-1}(\theta) \exp(-\mathcal{S}(X^{(k)}, X(\theta)))$ is computed using a scoring rule $\mathcal{S}$ based on real interaction data $X$ and model simulations $X(\theta)$. SMC is used to draw posterior samples ${\theta_1, \ldots, \theta_n}$, which generate simulated trajectories from $\mathcal{M}(\theta)$. An optimal policy is then trained by maximizing the expected action value over these $n$ trajectories, and the updated policy $\mu^{(k+1)}$ is used in the next episode.}
    \label{fig:enter-label}
\end{figure}

In Section \ref{sec:BRL} we provide an overview of model-based Bayesian RL and the relevance of Thompson sampling. In Section \ref{sec:exp_TS} we introduce expected Thompson sampling (ETS) and demonstrate its application on a simple “chain-task” example for which the likelihood function is available. Section \ref{sec:gbrl} details the generalized Bayesian inference framework, with Section \ref{sec:preqsr_post} discussing the properties of the prequential scoring rule posterior and Section \ref{sec:smc} explaining the SMC sampler used to sample from the posterior. In Section \ref{sec:etsq}, we derive an error bound for the approximate action-value function under the ETS framework. Simulation studies are presented in Section \ref{sec:results} to evaluate the approach, and Section \ref{sec:conclusion} summarizes key findings and conclusions. All proofs of the lemmas and theorems, as well as implementation details of the algorithms, are provided in the supplementary material.

\section{Model-based Bayesian Reinforcement learning}

\label{sec:BRL}

In RL, an agent (eg. a video game player) interacts with an unknown environment (eg. the virtual reality inside the game) by taking actions that cause the transition of the environment to a new state, yielding rewards in the process. Suppose, at time $t$, the environment was observed at the state $s_t \in \mathcal{S}$, where $\mathcal{S}$ is the state space. After the agent's action $a_t \in \mathcal{A}$ (the action space $\mathcal{A}$), the environment moves to the next state, $s_{t+1} \in \mathcal{S}$, and the agent receives a reward $r_{t+1} \in \mathcal{R}$ where $\mathcal{R}$ is the reward space. The goal of the agent is to devise a strategy that selects actions based on the current state to maximize cumulative rewards. Under the assumption that the new state ($s_{t+1}$) only depends upon the previous state and action ($s_t, a_t$) and not on $\lbrace s_u, a_u: u< t\rbrace$ (Markovian assumption), the environment can be considered as a Markov decision process (MDP), as defined below.

A {\em Markov decision process} (MDP) $\mathbb{M}$ on the state space $\mathcal{S}$ and action space $\mathcal{A}$, can be defined by the distribution of the initial state $\rho$ and the transition probabilities $P(S_{t+1} = s_{t+1} | S_t = s_t, A_t = a_t)$\footnote{Here, uppercase letters denote random variables, with their lowercase versions representing their realizations.} and $P(R_{t+1} = r_{t+1} | S_t = s_t, A_t = a_t)$. The ultimate objective is to develop a policy $\mu$ that maximizes the total reward over the long term in the future, where $\mu: \mathcal{S} \rightarrow \mathcal{A}$ is a mapping of the state space to the action space. This involves making decisions—choosing actions based on the current state—to navigate the environment in a way that accrues the maximum possible reward. Consequently, it becomes an optimization problem where the objective function is dependent on future rewards. 
To formalize this, we derive the value of a policy $\mu$ as the expected discounted return from any initial state if the policy $\mu$ is followed thereafter to interact with the environment. Hence, for any time $t$, we define the discounted return as $G_t = \sum_{k=0}^{\infty} \gamma^k {R}_{t+k+1}$. Then the value function of a policy $\mu$ at the state $S_t = s$, is defined as, 
$$
V^{\mathbb{M}}_{\mu} (s)  = \mathbb{E}_{\mathbb{M} , \mu} \left[ G_t| S_t = s\right] \quad \text{for all } s\in \mathcal{S}
$$
Here $\gamma \in (0, 1]$ is a discounting factor. A policy $\mu$ is said to be optimal for $\mathbb{M}$ if, $V_{\mu}^{\mathbb{M}}(s)=$ $\underset{\mu'}{\operatorname{max}} V_{\mu^{\prime}}^{\mathbb{M}}(s)$ for all $s \in \mathcal{S}$. 

However, as is often the case, instead of directly calculating the value function of a policy, we calculate the action-value function.  For any time $t$, the action-value function of a policy $\mu$ for the state-action pair, $(S_t = s, A_t = a)$ is defined as, 
$$
Q^{\mathbb{M}}_{\mu} (s, a)  = \mathbb{E}_{\mathbb{M} , \mu} \left[G_t | S_t = s, A_t = a\right] \quad \text{for all } s\in \mathcal{S}, a \in \mathcal{A}
$$

A natural way to assess the performance of a policy is via its \textbf{regret}. Regret quantifies the performance gap between a learned policy and the optimal policy, showing how much is lost due to suboptimal decisions. Let us denote the optimal policy under the MDP $\mathbb{M}$ by $\mu_0$. Then, the regret of a policy $\mu$  is given by, $\text {Regret}(\mu)= \sum_{s \in \mathcal{S}} \rho(s) (V^{\mathbb{M}}_{\mu_0} (s) - V^{\mathbb{M}}_{\mu} (s))$



\paragraph{Parametric MDP} Calculation of the value function is not straightforward since it involves taking expectations with respect to the MDP, which is not known in advance. Common approaches include using Bellman equations to approximate the value function based on observations from the environment received through interactions. These approaches are often sample-inefficient, as a precise approximation of expectations requires extensive training data. To mitigate this sample inefficiency, in \emph{model based RL} a parametric model for the MDP is assumed. If we can learn a good model emulating the true MDP, this can solve the above issue as we can draw many samples from the parametric MDP from which the optimal policy can be learned. But this efficiency can only be achieved if we can learn a good model even with smaller training datasets, which we answer positively here.  A parametric MDP $\mathbb{M}_{\theta}$, parameterized by $\theta = (\theta_{st}, \theta_{r}) \in \Theta$, is defined over the state space, $\mathcal{S}$ and action space $\mathcal{A}$. We further assume the action space be finite, meaning $\mathcal{A}$ is countable and $|\mathcal{A}| = n_a < \infty$.\footnote{Our proposed method and theoretical work will be developed under this assumption, but we will show empirically an extension of this for continuous action space, leaving the theoretical development as future work.} $\mathbb{M}_{\theta}$ can be described by the distribution of the initial state $\rho$ and  
\begin{align*}
    S_{t+1} | S_t = s_t, A_t = a_t &\sim P_{\theta_{st}} (.| s_t,a_t)\\
    R_{t+1} | S_t = s_t, A_t = a_t &\sim P_{\theta_{r}} (.| s_t,a_t)
\end{align*}
The distributions $P_{\theta_{st}}$ and $ P_{\theta_r}$ parameterized by $\theta_{st}$ and $\theta_r$ model the state transitions and reward distributions, respectively, for $\mathbb{M}_{\theta}$. For simplicity, the value function and the action-value function (or, the $Q$-function) of a policy $\mu$ under $\mathbb{M}_{\theta}$ will be denoted as $V^\theta_{\mu}$ and $Q^\theta_{\mu}$ respectively. Let $\mathcal{M} = \{\mathbb{M}_{\theta}: \theta \in \Theta \}$ represent a class of Markov decision processes parameterized by $\theta$. If the true underlying MDP, denoted by $\mathbb{M}_{0}$, does or does not belong to the model class $\mathcal{M}$, then we call $\mathbb{M}_{0}$ being well-specified or misspecified respectively.

\paragraph{Bayesian inference of parametric MDP}Suppose we have generated a trajectory of state-action-reward-next state sequences of length $T$ through interacting with the unknown environment $\mathbb{M}$, given by $H_T = {h}_T$ where $h_T = \{(s_t, a_t, r_{t+1}, s_{t+1})\}_{t=1}^{T}$. To learn $\mathbb{M}_{\theta}$ or the value of the parameters $\theta$, we first define the log likelihood function for $\theta$ with respect to the observed trajectory under $\mathbb{M}_{\theta}$ as,
\begin{eqnarray}
\label{eq:loglik_MDP}
    L_{T}(\theta) = \sum_{t=1}^{T} \log P_{\theta_{st}}(s_{t+1}| s_t, a_t) +  \log P_{\theta_{r}}(r_{t+1}| s_t, a_t).
\end{eqnarray}
We can update prior beliefs about the model parameters with the likelihood function using the Bayes theorem
\begin{equation}\label{bayesth}
   \log \pi(\theta | {h}_T )  \propto \log \pi(\theta)  + L_{T}(\theta) 
\end{equation}
where the prior distribution $\pi(\theta)$ quantifies our prior belief and posterior distribution $\pi(\theta | {h}_T )$ is our updated belief. 

\emph{model-based Bayesian RL} uses the posterior distribution of parametric MDP $\mathbb{M}_{\theta}$ learned via Bayesian inference, to learn an optimal policy under unknown uncertain environments. We first introduce Thompson sampling, a popular approach that only uses one parametric MDP sampled from the posterior distribution, and then propose expected Thompson sampling in Section~\ref{sec:exp_TS} which instead uses many independent and identically drawn parametric MDPs from the posterior distribution.

\paragraph{Thompson sampling (TS)} Thompson sampling (TS)\citep{thompson1933likelihood} is a widely used approach in multi-armed bandit (MAB) problems \citep{lattimore2020bandit}. In a typical MAB setup, each arm (or action) is associated with a reward, drawn from an unknown distribution. The agent’s objective is to maximize the accumulated reward through interactions with this environment. To gain insight into the reward distribution, the agent must experiment by trying different arms.

A Bayesian solution to this challenge begins with a prior belief in the mean rewards, representing our initial knowledge about the environment. As the agent interacts with the environment, it collects a sequence of action-reward pairs. This information is used to update the prior, refining the posterior distribution of the rewards. The posterior distribution is used to design a policy based on this updated knowledge.

In TS, the agent draws a sample from the updated posterior of the reward distribution and chooses the arm associated with the highest sampled reward to interact with the environment in the next round. Thus, TS offers an effective method to address the exploration-exploitation dilemma in MAB problems. This approach has been generalized to reinforcement learning (RL) tasks as well \citep{gopalan2015thompson,ouyang2017learning}. In RL contexts, alongside rewards, the agent observes the environment's state. In this case, the posterior distribution of the MDP parameters, $\theta$ becomes the quantity of interest.

\subsection{Expected Thompson sampling}
\label{sec:exp_TS}
Thompson Sampling (TS) balances exploration and exploitation by updating a policy based on a sample from the posterior distribution of the model parameters. However, relying on a single posterior sample may introduce excessive noise, making it difficult to fully exploit the information within the posterior. A more stable and reliable approach would be to estimate the value function using multiple posterior samples. As the number of samples increases, the standard error decreases, leading to a more accurate estimate. Building on this idea, we introduce the expected Thompson sampling (ETS) algorithm, an extension of TS that leverages multiple samples from the posterior to provide a more reliable estimate of the underlying value function.

To reduce computational overhead, we update the posterior after a fixed number of interactions rather than after every step. In RL, episodic tasks—with clear terminal states—naturally allow for policy updates at the episode end. For non-episodic (infinite-horizon) tasks, we define episodes of fixed length $\tau$ and update the policy accordingly. While we assume constant episode length for simplicity, our results can be extended to variable lengths. We now outline our episodic policy update strategy using ETS.

In settings with a finite, discrete action space $\mathcal{A}$, the optimal policy is obtained by maximizing the $Q$-function. Let, $\pi_k(\theta) = \pi(\theta \mid h_{\tau k})$ denote the posterior distribution over model parameters after the $k$-th episode. In TS-based methods \citep{dimitrakakis2013abc}, a single parameter sample $\theta_k \sim \pi_k$ is drawn, and the policy $\mu$ is optimized based on the estimated $Q$-function $Q_{\mu}^{\theta_k}$ from simulated trajectories using the model $\mathbb{M}_{\theta_k}$.

In contrast, we propose to use multiple posterior samples. Let $\bm{\theta}^{(k)} = \{\theta_{ki}\}_{i=1}^n$ represents a set of $n$ posterior samples from $\pi_k$. The $Q$-function for policy $\mu$ under ETS is then estimated by:
\begin{equation}\label{eq:avg_q}
    Q^{\bm{\theta}^{(k)}}_{\mu}  = \frac{1}{n} \sum_{i=1}^n Q^{{\theta}_{ki}}_{\mu} \approx \int_{\Theta} Q^{{\theta}}_{\mu} \pi_k(\theta) d\theta, \quad \text{for all $(s,a) \in \mathcal{S} \times \mathcal{A}$}.
\end{equation}

\paragraph{Policy iteration} Policy iteration methods are a widely used class of algorithms for solving problems with discrete state spaces. These methods begin with an initial policy—often chosen at random—and then evaluate its performance by calculating the corresponding Q-function, a process known as the policy evaluation step. Following this, a new and improved policy is generated by selecting actions that maximize the estimated Q-function; this is referred to as policy improvement. The agent alternates between these two steps, iterating until the policy converges, at which point the algorithm reaches a solution.

ETS can be easily integrated with policy iteration methods by performing the policy evaluation using equation (\ref{eq:avg_q}). In the following, we provide pseudocode as Algorithm \ref{alg1}.

\begin{algorithm}
\caption{Policy iteration using expected Thompson sampling}
\label{alg1}
\begin{algorithmic}[]
\State \textbf{Input: } Prior distribution, $\pi$
    \State Observe $h_{\tau} = \{(s_t, a_t, r_{t+1}, s_{t+1})\}_{t=1}^{\tau}$ by playing a random policy ($ = \mu^{(1)}$, say )
    \State Update the posterior, $ \log \pi_1(\theta) \propto \log \pi(\theta) +  L_{\tau}(\theta)$
    \For{\textit{episodes $k =1, 2, \ldots$}}
        \State Sample $\bm{\theta}^{(k)}= \{\theta_{ki}\}_{i=1}^n \sim \pi_k(\cdot) $
        \State Consider the initial policy $\mu_0 = \mu^{(k)}$
        \For {$j =1, 2, \ldots J$}
            \State Compute $Q^{\bm{\theta}^{(k)}}_{\mu_{j-1}} = \frac{1}{n} \sum_{i=1}^n Q^{{\theta}_{ki}}_{\mu_{j-1}}$
            \State Update the policy $\mu_j (s) = \arg \max_{a} Q^{\bm{\theta}^{(k)}}_{\mu_{j-1}}(s, a)$
            \EndFor
        \State Set the policy for next episode: $\mu^{(k + 1)} = \mu_J$
        \For{\textit{timesteps $t =1,2, \ldots,\tau$}}
            \State Play $a_{\tau (k +1) +t} = \mu^{(k + 1)}(s_{\tau (k + 1)+t}) $
            \State Observe $r_{\tau (k + 1)+t+1}$ and $s_{\tau (k+1)+t+1}$
        \EndFor
    \EndFor
\end{algorithmic}
\end{algorithm}

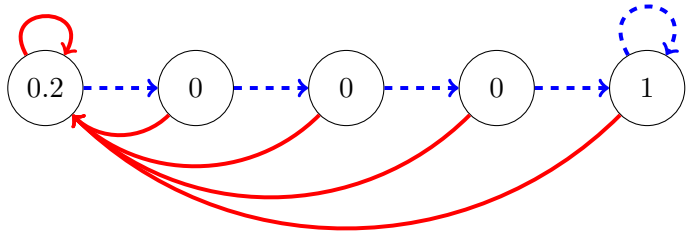
\begin{figure}
    \centering
    \begin{tikzpicture}
    
        \foreach \i/\num in {1/0.2,2/0,3/0,4/0,5/1}
            \node[circle, draw, minimum size=1cm] (C\i) at (\i*2,0) {$\num$};
        
        \foreach \i in {1,...,4}
            \draw[blue, dashed, ->, line width=1.5pt] (C\i) -- (C\the\numexpr\i+1\relax);
        
        \foreach \i in {2,...,5}
            \draw[red, ->, bend left=45, line width=1.5pt] (C\i) to (C1);
        
        \draw[red, ->, line width=1.5pt] (C1) to [out=120, in=60, looseness=4](C1);
        
        \draw[blue, dashed, ->, line width=1.5pt] (C5) to [out=120, in=60, looseness=5] (C5);
    \end{tikzpicture}
    \caption{The chain task (Adapted from Figure 7.3 in \citet{dimitrakakis2022decision})}
    \label{fig:chain}
\end{figure}

\paragraph{Example}
To demonstrate the effectiveness of the expected Thompson Sampling algorithm, here we use the `chain task' from \citet{dimitrakakis2022decision}. The task has two actions and five states, as shown in Fig.(\ref{fig:chain}). The task always starts from the leftmost state ($s(1)$, say) where the mean reward is $0.2$. There are no rewards assigned to the intermediate states. The mean reward at the terminal state (rightmost state) is 1. The first action which is denoted by the dashed-blue line takes the agent to the right, whilst the other action denoted by the red-solid line takes the agent to the first state. However, there is a probability $0.2$ that actions act in a reverse way in the environment.

The chain task is a very simple task that has the typical exploration-exploitation dilemma. Although there is a small reward at the first state, there is a bigger reward for reaching the last state. For a fairly big horizon, taking the right action is the optimal policy that we want our algorithm to learn. 

Note that, here, the true model for the environment's dynamics can be expressed with $Binomial (p_0=0.8)$ transition probability and rewards distributed as $\mathcal{N}(\overline{r}_0, I_5)$, with $\overline{r}_0 = (0.2, 0, 0, 0, 1)$ for the 5 states. Here we assume the parametric model ($\mathbb{M}_{\theta}$) for the underlying MDP is known to us except for the true model parameters, $\theta_0 = (p_0, \overline{r}_0)$. Since the true underlying MDP $\mathbb{M}_0 = \mathbb{M}_{\theta_0}$ belongs to the class $\mathbb{M}_{\theta}$, this is an example of well-specified model. To infer the model parameters, we started by assigning a Beta prior to the transition probabilities and a Gaussian prior to the mean rewards. Using the conjugacy of the prior distributions, we drew samples from the exact conjugate posterior distribution to perform ETS.

For policy learning, we have used a dynamic programming algorithm called backward induction (BI) following the implementation of \citet{dimitrakakis2022decision}. This method is well-suited for a finite-horizon RL task with finite state and action space. The BI algorithm uses Bellman's equations to derive the action value function starting from the terminal state and going backward to calculate it for each state recursively. To accommodate ETS, within each episode of interaction, for each state, we calculate the $Q$ function based on all the sampled MDPs according to equation (\ref{eq:avg_q}) and choose the greedy action that maximizes the pooled estimate of the $Q$ function.

\begin{figure}
     \centering
     \begin{subfigure}[t]{0.48\linewidth}
         \centering
         \includegraphics[width=\linewidth]{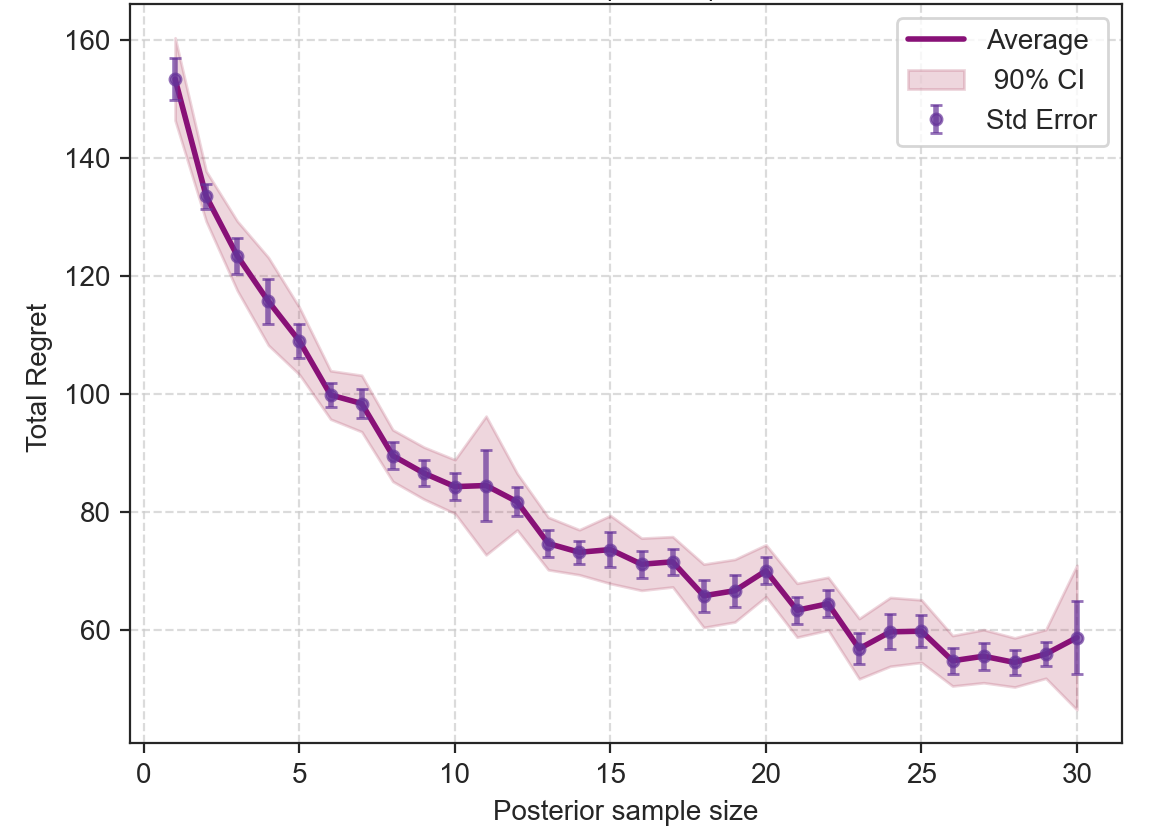}
         \caption{Plot of total regret for different posterior sample size}
         \label{total_regret}
     \end{subfigure}
     \hfill
     \begin{subfigure}[t]{0.49\linewidth}
         \centering
         \includegraphics[width=\linewidth]{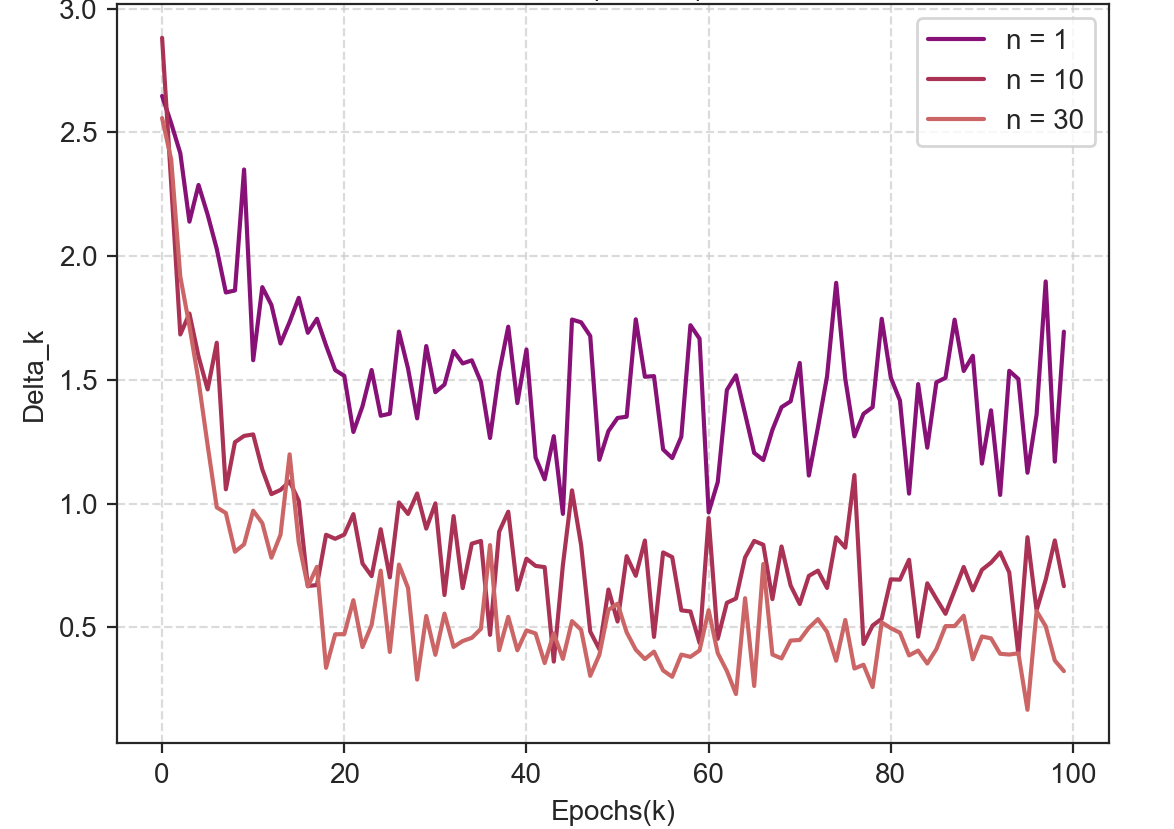}
         \caption{Plot of episodic regrets over the episodes}
         \label{deltak}
     \end{subfigure}
     \caption{The chain experiment was conducted for  $K = 100$ episodes with an episode length of $\tau = 20$. The same experiment was repeated 30 times, with the average values plotted in these graphs.}
     \label{bigfig}
\end{figure}

Now let $\mu_0$ be the optimal policy under the true MDP which is unknown to the agent. Suppose the agent interacts with the environment over epochs or episodes, $k = 1, 2, \ldots$ and ${\mu}^{(k)}$ is the policy to determine action from a given state during the $k$th episode using ETS. The total regret of ETS due to using the sequence of policies ${\mu}^{(1)}, {\mu}^{(2)}, \dots$ over a total of $T$ timesteps is defined as the sum of episodic regrets as shown below,

\begin{align*}
    \text { Regret }(T) &=\sum_{k=1}^{[ T / \tau]} \sum_{s \in \mathcal{S}} \rho(s) \Delta_{k, s} \quad \text{ with } \Delta_{k, s} =  V^{\mathbb{M}_0}_{\mu_0} (s) - V^{\mathbb{M}_0}_{{\mu}^{(k)}} (s)\\
    &= \sum_{k=1}^{[ T / \tau]} \Delta_{k, s(1)},
\end{align*}
since an episode always starts from the leftmost state, $s(1)$.

 Fig. (\ref{total_regret}) illustrates that total regret decreases as we increase the number of samples from the posterior to evaluate the action value functions to learn the optimal policy. Similarly, in Fig. (\ref{deltak}), we observe that the sequence of episodic regrets $\{\Delta_{k, s(1)}\}$ vanishes more rapidly as we incorporate more samples to estimate the value function, thus reducing the estimation variability that supports the statement of Theorem \ref{th:regret}.

Hence, these figures demonstrate that learning the optimal policy can be accelerated by taking larger samples from the posterior to estimate the value functions. Consequently, the ETS algorithm outperforms standard Thompson sampling (TS) in terms of lower regret. The ETS method operates on the principle that utilizing multiple samples from the posterior can yield a more stable approximation of the value function. Despite this advantage, traditional TS has been more widely adopted because of its lower implementation cost, even though it may suffer from approximation errors.








\section{(Generalized) Bayesian model-based RL}
\label{sec:gbrl}
Bayesian model-based RL requires a Bayesian treatment of the model parameters. However, when dealing with complex models with intractable likelihoods (e.g. when the analytical form for $P_{\theta_{st}}$ and $ P_{\theta_r}$ is unavailable for a parametric MDP), directly calculating the posterior distribution becomes challenging. 
Nevertheless, for most of the assumed parametric MDP models, we can directly sample from them, a distinctive feature that we capitalize upon. This capability to derive samples directly from the model suggests using a likelihood-free inference framework. This approach entails generating data by running simulations based on the model parameters. The samples we generate from the simulator contain a spectrum of potential outcomes under the model, given the population parameters. 
By using methods available under the broad umbrella of likelihood-free inference (LFI), these samples let us make inferences without having to confront the mathematical intractability of the likelihood function. Two of the most popular classes of LFI methods are approximate Bayesian computation (ABC) \citep{lintusaari2017fundamentals} and Bayesian synthetic likelihood (BSL) \citep{price2018bayesian}, of which ABC has been used before for Thompson Sampling in \citep{dimitrakakis2013abc}. They approximate the intractable likelihood function either implicitly or explicitly using these samples. The asymptotic contraction of these approximate posteriors towards the true parameter value depends upon the choice of the summary statistics and some conditions being satisfied by the chosen summary statistics \citep{frazier2018asymptotic, li2018asymptotic, frazier2023bayesian}, which are difficult to verify in practice.

    In this work, we choose a different likelihood approximation method, the scoring rule posterior framework of \cite{pacchiardi2024generalized} which facilitates likelihood-free inference using a suitable scoring rule for a given type of data. This is similar to BSL in the sense that it provides an explicit approximation to the likelihood but it has outperformed both BSL and ABC regarding computational efficiency specifically more for high-dimensional examples. Further, it is easy to verify that the scoring rule posterior contracts to the true parameter value when a strictly proper scoring rule is chosen \citep{pacchiardi2024generalized}. Next, we explain the scoring rule posterior and extend them for the inference of parametric MDPs, by using scoring rule posteriors based on prequential scoring rules as introduced in \cite{pacchiardi2024probabilistic}. 

\subsection{Scoring rule posterior}
\label{sec:sr_post}
For models with intractable likelihood function $p(\mathbf{y}^{obs} \mid \theta)$, the posterior distribution cannot be computed directly via Bayes’ theorem:
$$
\begin{aligned}
\pi(\theta \mid \mathbf{y}^{obs}) \propto \pi(\theta) p( \mathbf{y}^{obs}\mid \theta) =  \pi(\theta) \exp \left\{ \log p(\mathbf{y}^{obs} \mid \theta) \right\}, 
\end{aligned}
$$
where $\theta$ is the model parameter and $\pi(\theta)$ is a prior distribution on the parameter space $\Theta$. To mitigate this, \cite{pacchiardi2024generalized} consider loss functions $S(P_{\theta},\mathbf{y})$ known as scoring rules \citep{gneiting2007strictly} that measure the fit between the distribution $P_{\theta}$ of the data under parameter $\theta$, and an observed data point $\mathbf{y}$. These loss functions allow inference without access to the likelihood, requiring only the ability to simulate from $P_{\theta}$.

The likelihood-free \textit{scoring rule posterior} is then defined as follows:
$$
\pi_{S}\left(\theta \mid \mathbf{y}^{obs}\right) \propto \pi(\theta) \exp \left\{-w S\left(P_{\theta}, \mathbf{y}^{obs}\right)\right\}.
$$
Comparing the two expressions, we note that the (negative) log-likelihood function can itself be considered a scoring rule (known as log-score, \citep{Dawid_2014}) and that we have introduced an additional parameter $w$, which is known as the learning rate in generalized Bayesian inference \citep{holmes2017assigning} controlling the relative weighting of the observations relative to the prior. 
Next, we discuss some properties of different types of scoring rules that ensure asymptotic contraction of the resulting posterior distributions. 

Suppose $P_0$ is the underlying true data-generating process. Then, we can define the expected scoring rule as, $\tilde{S}(P_{\theta}, P_0) = \mathbb{E}_{Y \sim P_0} S(P_{\theta}, Y)$. A scoring rule $S$ is said to be \textit{proper} if $\tilde{S}(P_{\theta}, P_0)$ is minimized when the assumed distribution $P_\theta$ is equal to the distribution $P_0$ generating the observed data $\mathbf{y}^{obs}$. If $P_\theta = P_0$ is the unique minimum, the scoring rule is said to be \textit{strictly proper}. Thus, when $S$ is strictly proper we can define a statistical divergence between the distributions $P_{\theta}$ and $P_0$ as:
\begin{align}
    D(P_{\theta}, P_0) = \tilde{S}(P_{\theta}, P_0) - \tilde{S}(P_0, P_0)
\end{align}
This divergence is non-negative and equals zero if and only if $P_{\theta} = P_0$. This is a generalized divergence that measures how well the model $P_{\theta}$ approximates the true data-generating process $P_0$. If we take the scoring rule to be negative log-likelihood or the log score, then this divergence is equivalent to the Kullback-Leibler divergence. 

Some examples of scoring rules are the Continuous Ranked Probability Score (CRPS) \citep{szekely2005new}, Energy score or Kernel score \citep{gneiting2007strictly} etc. For later sections, we will be using the energy score, which can be seen as a multivariate generalization of CRPS, defined as, 
    \begin{equation}\label{esr}
    S_E^{(\beta)}(P_{\theta}, \mathbf{y}) =2 \cdot \mathbb{E} || X - \mathbf{y}||^\beta - \mathbb{E} ||X- X'||^\beta, \quad X,  X' \sim P_{\theta};\quad \beta \in (0,2)
    \end{equation}
This is a strictly proper scoring rule for the class of probability measures 
$\mathcal{P} = \{P_{\theta} : \mathbb{E}_{X \sim P_{\theta}} ||X||^{\beta} < \infty, \forall \theta \in \Theta\}$, when $\beta \in (0,2)$ \citep{gneiting2007strictly}. The related divergence is the square of the energy distance, which is a metric between probability distributions. Further, the energy score can be unbiasedly estimated using $\mathbf{x}_j \sim P_{\theta}, j=1, \ldots, m$ which are independent and identically distributed samples from the model $P_{\theta}$. The unbiased estimate for the energy score in equation (\ref{esr}) can be obtained by Monte Carlo estimates of the expectations in $S_{\mathrm{E}}^{(\beta)}(P, y)$
    \begin{equation}\label{mc-esr}
    \hat{S}_{\mathrm{E}}^{(\beta)}\left(\left\{x_j\right\}_{j=1}^m, \mathbf{y}\right)=\frac{2}{m} \sum_{j=1}^m\left\|x_j-\mathbf{y}\right\|^\beta-\frac{1}{m(m-1)} \sum_{\substack{j, k=1 \\ k \neq j}}^m\left\|x_j-x_k\right\|^\beta \quad,  \beta \in (0,2).
    \end{equation}
For our implementation, we will consider $\beta = 1$ and write $S_{\mathrm{E}}^{(1)}(P, \mathbf{y})$ simply as $S_{\mathrm{E}}(P, \mathbf{y})$.

Theoretical properties of asymptotic normality and generalization bound for scoring rule posteriors have been studied in \cite{giummole2019objective,  pacchiardi2022statistical, pacchiardi2024generalized}. In particular, when the scoring rule is strictly proper, the corresponding posterior contracts around the true model parameter when the model is well-specified. Additionally, the scoring rule posterior for some scoring rules (e.g. energy score or kernel score) exhibits robustness against outliers compared to the standard Bayes posterior using log-score (more details in Chapter 3 of \cite{pacchiardi2022statistical}).

\subsection{Prequential scoring rule posterior}
\label{sec:preqsr_post}
For complex simulator models--such as deep generative models often used in R; we typically lack closed-form expressions for the conditional distributions, $P_{\theta_{st}}$ and $ P_{\theta_r}$, which define the log-likelihood of the Markov process (see equation (\ref{eq:loglik_MDP})). However, such models can still generate simulations efficiently. Hence combining the idea of scoring rule posterior explained in the previous section with the idea of \emph{prequential (predictive sequential) SR} proposed in \cite{pacchiardi2024probabilistic}, here we propose \emph{prequential scoring rule posterior}. As before, we use Scoring Rules (SRs) to assess the goodness of fit of the one-step-ahead predictive distribution, conditioned on the previously observed value, to the current observation. By summing a sequence of observations over a period of time, the cumulative SR evaluates the predictive performance of a sequence of conditional models. This cumulative measure is referred to as the prequential score. Thus, on observing the trajectory $H_T = h_T$, the prequential SR for a parametric MDP ($\mathbb{M}_{\theta}$) based on a scoring rule $S$ can be defined as, 
\begin{equation}\label{preq}
    \PS(\mathbb{M}_{\theta}, h_T) = \sum_{t=1}^{T} \left(S({P}_{\theta_{st}}(.|s_{t}, a_{t}), s_{t+1}) + S({P}_{\theta_{r}}(.|s_{t}, a_{t}), r_{t+1}) \right).
\end{equation}
We notice if we have access to the analytical form of $P_{\theta_{st}}$ and $ P_{\theta_r}$, then taking $S$ to be the log-score the above equation reduces to the log-likelihood defined in equation (\ref{eq:loglik_MDP}). Moreover, when the scoring rule $S$ is (strictly) proper, the prequential scoring rule $\PS(\mathbb{M}_{\theta}, H_T)$ is also (strictly) proper for the class of all Markovian conditional distributions over the next state and reward given the previous state-action pair as shown in Theorem 2 of \cite{pacchiardi2024probabilistic}. Also, if we can estimate the SR $S$ as shown in equation (\ref{mc-esr}) unbiasedly, then the prequential SR $\PS$ can also be estimated unbiasedly by generating $m$ simulations of next states and reward until time $T$, conditioned on previously observed state and action.

Often in RL problems, the reward distribution is chosen deterministically and it is considered as a constant function of state and action. Hence, in those cases, our $\PS$ defined in equation (\ref{preq}) simplifies to 
\begin{equation}\label{eqn: psr}
\PS_T(\mathbb{M}_{\theta}, h_T) = \sum_{t=1}^{T} S({P}_{\theta}(.|s_{t}, a_{t}), s_{t+1}), 
\end{equation}
and $\theta$ only contains parameters $\theta_{st}$. Now suppose $\mathbb{M}_0$ is the true data generating process, then we can define the expected prequential score as, 
\begin{equation*}
    \widetilde{\PS}_T(\mathbb{M}_{\theta}, \mathbb{M}_0) =\mathbb{E}_{H_T \sim \mathbb{M}_0}\PS(\mathbb{M}_{\theta}, H_T).
\end{equation*}
From now on, we use the notations $\PS_T(\theta)$ and $\widetilde{\PS}_T(\theta)$ to denote the empirical prequential score $\PS_T(\mathbb{M}_{\theta}, h_T)$ and the expected prequential score $\widetilde{\PS}_T(\mathbb{M}_{\theta}, \mathbb{M}_0)$, respectively. Now without loss of generality, we redefine our prequential scoring rule posterior on the parameter $\theta$ of the parametric MDP as, 
\begin{eqnarray}\label{eq: gen_post_w}
\pi_{\PS}\left(\theta \mid h_T\right) \propto \pi(\theta) \exp \left\{-w \PS_T(\theta) \right\}. 
\end{eqnarray}
Note that choosing values of \( w \neq 1 \) can be interpreted as a form of annealing applied to the target posterior—\emph{heating} for \( w < 1 \) and \emph{cooling} for \( w > 1 \).\footnote{In the likelihood-based setting (i.e., using the log score), setting \( w \neq 1 \) yields power posteriors. The influence of $w$ on the asymptotic properties of such posteriors has been studied in \cite{Ray_2023}.} We fix \( w = 1 \) in our experiments, as our focus is on the long-term behavior of the posterior over longer sequences of trajectories. We now proceed to analyze the asymptotic properties of the generalized posterior defined above.


\paragraph{Asymptotic properties of the generalized prequential posterior:} In this section we derive the asymptotic properties of the generalized prequential posterior. We will prove a Bernstein--von Mises (BvM) theorem under some assumptions. Before stating the assumptions, we first clarify some mathematical notations. We denote the parameter space as $\Theta \subseteq \mathbb{R}^p$. We denote the gradient, the matrix of second-order derivatives, and the vector of third-order derivatives of a function $f(\theta)$ with respect to $\theta$ as
$f'(\theta) = \left(\frac{\partial f(\theta)} {\partial \theta^i} (\theta)\right)_{i=1}^p \in \mathbb{R}^p$, 
$f''(\theta) = \left(\frac{\partial f}{\partial \theta^i \partial \theta^j}(\theta)\right)_{i, j = 1}^p \in \mathbb{R}^{p \times p}$ and $f'''(\theta) = \left(\frac{\partial f}{\partial \theta^i \partial \theta^j \partial \theta_k} (\theta)\right)_{i, j, k = 1}^p \in \mathbb{R}^{p^3}$ respectively. For a given $\theta \in \mathbb{R}^p$ and $r>0$, we would denote the open ball of radius $r$ around $\theta$ as $B_{r}(\theta) = \{\theta' \in \mathbb{R}^p: ||\theta' - \theta|| < r\}$, where $||\cdot||$ stands for Euclidean norm.

We begin by analyzing the limiting behaviour of the expected prequential score as we observe longer trajectories of interaction with the environment under the following assumption.

\begin{enumerate}[label=\textbf{A\arabic*}]
    \setcounter{enumi}{0}

    \item \label{assumption1} $S$ is a strictly proper scoring rule and, the time-averaged generalized entropy of the true model $\frac{1}{T} \widetilde{\PS}_T(\mathbb{M}_0, \mathbb{M}_0)$ has a finite limit as $T \to \infty$.   
\end{enumerate}

\begin{lemma}\label{lem-limiting_exp-loss}
Under Assumption~\ref{assumption1}, there exists a function \( \PS^*(\theta) \) such that as \( T \to \infty \), \( \frac{1}{T}\widetilde{\PS}_T(\theta) \to \PS^*(\theta) \) uniformly with probability one under $\mathbb{M}_0$. 
\end{lemma}
An outline of the proof of the Lemma \ref{lem-limiting_exp-loss} is provided in Appendix A.1.
Next, let us denote the minimizer of the (normalized) empirical prequential score as 
\begin{equation*}
    \hat{\theta}_T = \arg \min_{\theta \in \Theta} \frac{1}{T}\PS_T(\theta).
\end{equation*}
 A key step in establishing the consistency of this estimator is a uniform law of large numbers (ULLN) for the empirical prequential scores. However, classical ULLNs assume i.i.d. observations, which do not apply here as the data are generated by a Markov process. We therefore impose mixing conditions and exploit an action-based decomposition of the prequential score to obtain a ULLN.

Now suppose the action space $\mathcal{A}$ is finite. Then, we can decompose the prequential score as, 
$$
\mathcal{PS}_T(\theta) = \sum_{a \in \mathcal{A}} \mathcal{PS}_T^a(\theta),
$$
where $\mathcal{PS}_T^a(\theta) = \PS_T(\mathbb{M}_{\theta}, h_T^a)= \sum_{t=1}^{T} S({P}_{\theta}(.|s_{t}, a_{t}), s_{t+1}) \mathbb{I}_{\{a_t = a\}} $ records the scoring rule only for the time-steps at which action $a$ was taken upto time $T$. In addition, $T_a$ denotes the set of such time points, i.e. $T_a = \{t \leq T: a_t = a\}$ for each $a \in \mathcal{A}$. So, on fixing the current action, the simulator model $P_{\theta}$ merely predicts $s_{t+1}$ conditioned on the current state $s_t$. We use this decomposition along with the following assumptions on the true data generating process $\mathbb{M}_0$ to establish a ULLN.

\begin{enumerate}[label=\textbf{A\arabic*}]
    \setcounter{enumi}{1}
    \item \label{assumption2} (Asymptotic stationarity) For all actions $a$ in the action space, let, $G_t^a$ be the marginal distribution of $(S_t, S_{t+1})$ when $A_t = a$ for $t \geq 1$; then, $\frac{1}{|T_a|} G_t^a$ converges weakly to some probability measure $G^a$ on $\mathcal{S}^2$ as $T \to \infty$. 

    \item \label{assumption3} For all actions $a$ in the action space, both of the conditions below are satisfied:
    \begin{enumerate}
        \item (Mixing) Suppose $(S_t)_t^a$ is the sequence of the states at which action $a$ is taken. Either one of the following conditions holds:
        \begin{enumerate}
            \item $(S_t)_t^a$ is $\alpha$-mixing with mixing coefficient of size $r/(2r-1)$, with $r \geq 1$, or
            \item $(S_t)_t^a$ is $\gamma$-mixing with mixing coefficient of size $r/(r-1)$, with $r > 1$.
        \end{enumerate}
        \item (Moment boundedness) Define $J^a(s_t, s_{t+1}) = \sup_{\theta \in \Theta} |S(P_{\theta}(\cdot | s_t, a), s_{t+1})|$, then
        $$
        \sup_{t \geq 1} \mathbb{E}[J^a(s_t, s_{t+1})^{r + \delta}] < \infty,
        $$
        for some $\delta >0$, for the value of $r$ corresponding to the condition above which is satisfied.
     \end{enumerate}
\end{enumerate}

\begin{lemma}\label{lem-ulln}
(Uniform law of large numbers) Under assumptions \ref{assumption2} and \ref{assumption3}, a ULLN holds, which implies with probability $1$ under $\mathbb{M}_0$,
    $$
    \sup_{\theta \in \Theta}  |\PS_T(\theta) - \widetilde{\PS}_T (\theta)| \to 0 \text{ as } T \to \infty.
    $$
\end{lemma}
Proof of Lemma \ref{lem-ulln} is provided in Appendix A.2. We use a result from \cite{potscher1989uniform} as adapted by \cite{pacchiardi2024probabilistic} to establish the ULLN for prequential scores under the true data-generating process, assuming $\mathbb{M}_0$ satisfies the asymptotic stationarity and mixing conditions stated in Assumptions~\ref{assumption2} and~\ref{assumption3}. Intuitively, asymptotic stationarity implies that the joint distribution of any two consecutive states, conditioned on an action, converges to a stationary distribution. The mixing conditions ensure that, under a fixed action, the dependence between states decays rapidly as their time separation increases. These properties automatically hold for the Markovian model $\mathbb{M}_0$, where, given an action, the next state depends only on the current state. However, we prove the result for a broader class of models which satisfy the conditions \ref{assumption2} and~\ref{assumption3}.

\begin{corollary}\label{cor}
    Under assumptions \ref{assumption1}-\ref{assumption3}, it follows from Lemma \ref{lem-limiting_exp-loss} and Lemma \ref{lem-ulln} that, 
    $$
    \sup_{\theta \in \Theta}  |\frac{1}{T}\PS_T(\theta) - {\PS}^* (\theta)| \to 0 \text{ as } T \to \infty,
    $$
    with probability $1$ under $\mathbb{M}_0$.
\end{corollary}

 The above corollary is a direct consequence of Lemma \ref{lem-limiting_exp-loss} and Lemma \ref{lem-ulln} and a short proof can be found in Appendix A.3. Next, we use this result to show the asymptotic consistency of the estimators $\{\hat{\theta}_T\}_{T=1}^{\infty}$ under the following assumptions. 

\begin{enumerate}[label=\textbf{A\arabic*}]
    \setcounter{enumi}{3} 
    

    \item \label{assumption4} The parameter space $\Theta \subseteq \mathbb{R}^p$ is compact.

    \item \label{assumption5} $\theta^*$ is a unique minimizer of $\PS ^*(\theta)$, and there exists a metric $d$ on $\Theta$ such that, for all $\epsilon >0, $
    $$
    \min_{\theta: d(\theta, \theta^*) \geq \epsilon} {\PS}^*(\theta) - {\PS}^*(\theta^*) > 0.
    $$


\end{enumerate}

\begin{lemma}\label{lem-consistent_estimators_star}
    (Asymptotic consistency) Under assumptions \ref{assumption1}-\ref{assumption5}, as $T \to \infty$, we have with probability $1$ under $\mathbb{M}_0$,
    $$
    d(\hat{\theta}_T, \theta^*) \to 0.
    $$
\end{lemma}
We derive the above lemma using a result from \cite{skouras1998optimal} and a detailed proof can be found in Appendix A.4. Here the parameter space needs to be compact and the empirical prequential score must be smooth enough in the neighbourhood of $\theta^*$ as in condition \ref{assumption5}. Then, this result is analogous to the consistency of maximum likelihood estimators (MLE). Specifically, when the scoring rule \( S \) is the log score, the estimator \( \hat{\theta}_T \) is the MLE. The result establishes that \( \hat{\theta}_T \) is consistent for \( \theta^* \), the true parameter under a well-specified model, i.e., when \( \mathbb{M}_0 = \mathbb{M}_{\theta^*} \). In the misspecified case, where \( \mathbb{M}_0 \notin \mathcal{M} \), the parameter \( \theta^* \) corresponds to the model within \( \mathcal{M} \) that minimizes the expected prequential loss—i.e., the model with optimal one-step-ahead predictive performance.
Using these consistent estimators of \( \theta^* \), we next state the BvM theorem for the generalized prequential posterior under the following assumptions.

\begin{enumerate}[label=\textbf{A\arabic*}]
    \setcounter{enumi}{5}
\item \label{assumption6} $\pi : \mathbb{R}^p \rightarrow \mathbb{R}$ is a probability distribution with respect to the Lebesgue measure such that $\pi$ is continuous at $\theta^*$ and $\pi(\theta^*) > 0 $.

\item \label{assumption7} $E \subseteq \mathbb{R}^p$ is open and convex and let $\theta^*, \hat{\theta}_T \in E$ for all $T$ sufficiently large.

 \item \label{assumption8} $\PS_T''(\theta^*) \to H^*$  as $T \to \infty$ for some positive definite $H^*$.

\item  \label{assumption9} $\PS_T(\theta)$ have continuous third derivatives in $E$ and the third derivatives $\PS^{'''}_T(\theta)$ are uniformly bounded in $E$.

    \item \label{assumption10}For any $\epsilon >0 $, $\lim \inf_T \inf_{\theta: d(\theta, \hat{\theta}_T) \geq \epsilon} (\PS_T(\theta) - \PS_T(\hat{\theta}_T)) >0.$
    
\end{enumerate}

\begin{theorem}
\label{th-bvm}
(Bernstein-von Mises theorem) Let us define the generalized prequential pospterior distribution as $\pi_{\PS}(\theta|H_T)=\exp \left(-\PS_T(\theta)\right) \pi(\theta) / z_T$, where $z_T$ is a normalizing constant given by,  $z_T=\int_{\mathbb{R}^d} \exp \left(-\PS_T(\theta)\right) \pi(\theta) d \theta$. Then, under assumptions \ref{assumption1}-\ref{assumption10}, we have,

    \begin{equation}\label{eqth11}
     \int_{B_{\varepsilon}\left(\theta^*\right)} \pi_{\PS}(\theta|h_T) d \theta \underset{T \rightarrow \infty}{\longrightarrow} 1 \text { for all } \varepsilon>0
    \end{equation}
    that is, $\pi_{\PS}(\theta|h_T)$ concentrates at $\theta^*$;
    \begin{equation}\label{eqth12}
     z_T \sim \frac{\exp \left(-\PS_T\left(\theta_T\right)\right) \pi\left(\theta^*\right)}{\left|\operatorname{det} H^*\right|^{1 / 2}}\left(\frac{2 \pi}{T}\right)^{p / 2}
    \end{equation}
    as $T \rightarrow \infty$ (Laplace approximation); and letting $q_T$ be the density of $\sqrt{T}\left(\theta-\hat{\theta}_T\right)$ when $\theta \sim \pi_{\PS}\left(\theta \mid h_T\right)$,
     
    \begin{eqnarray}\label{eqth13}
    \int_{\mathbb{R}^D}\left|q_T(\theta)-\mathcal{N}\left(\theta \mid 0, H^{*-1}\right)\right| d \theta {\longrightarrow} 0
    \end{eqnarray}
    as $T \rightarrow \infty$, that is, $q_T$ converges almost surely to $\mathcal{N}\left(0, H^{*-1}\right)$ in total variation.

\end{theorem}
We prove the above BvM theorem using two results from \cite{JMLR:v22:20-469}, and we provide a sketch of the proof in Appendix A.5. For this result to hold, the prior distribution must be continuous and positive at $\theta^*$. Both $\theta^*$ and the sequence of its consistent estimators $\hat{\theta}_T$ must lie in an open and convex subset of $\Theta$. Further, the assumptions \ref{assumption8} and \ref{assumption9} ensure a Taylor series expansion of the empirical prequential score $\PS_T(\theta)$. In addition, we assume that $\PS_T(\theta)$ is smooth in the neighbourhood of $\hat{\theta}_T$ as in condition \ref{assumption10}. Then, the theorem states that as more data is observed, the generalized prequential scoring rule posterior concentrates around the $\theta^*$. It also begins to resemble a Gaussian distribution. The covariance matrix of this Gaussian is given by the inverse of $H^*$ which is a limit of $\PS''_T(\theta)$. Hence, $H^*$ plays the role of the Fisher information matrix in the likelihood-based framework.

\subsection{Sequential Monte Carlo with gradient-based kernel}
\label{sec:smc}
 
To sample from the prequential scoring rule posterior, here we propose to use a sequential Monte Carlo \citep{del2006sequential} scheme after every (or some) episode of interaction with the environment.

\paragraph{Sequential Monte Carlo (SMC)} 
Sequential Monte Carlo (SMC) refers to a class of algorithms that aim to represent a probability distribution through a set of weighted particles. Let $\{\pi_k\}$, for $k = 1, 2, \ldots$, to be the sequence of scoring rule posteriors using data from all the episodes up to the $k$-th episode. For each $k$, we use SMC to sample from the corresponding $\pi_k$, initializing the SMC with the posterior samples generated from the scoring rule posterior of the $(k-1)$-th episode. Hence the prior distribution used at the $k$-th episode to define the scoring rule posterior is $\pi_{k-1}$. To ensure a smooth transition from the prior ($\pi_{k-1}$) to the target distribution ($\pi_{k}$) at each episode, we introduce a series of intermediate target distributions. These distributions follow a geometric path, defined as ${\pi}^{(l)}_k(\theta) \propto \pi^{\alpha_{k,l}}_k(\theta) \pi_{k-1}^{1-\alpha_{k,l}}(\theta)$ with $0 \leq \alpha_{k,1} < \ldots \alpha_{k,L}=1$ as proposed by \citet{gelman1998simulating}. The sequence of temperatures $\{\alpha_{k,l}\}_{l=1}^L$ at the $k$-th episode can be determined adaptively based on effective sample size (ESS) \citep{beskos2015sequential} or conditional effective sample size (CESS) \citep{zhou2016toward}. 

An SMC sampler uses sequential importance sampling with a resampling approach on the sequence of targets. Starting from an initial set of weighted particles drawn from a proposal distribution, each SMC iteration propagates these particles toward the target distribution via a forward kernel. The resulting trajectories are reweighted. The particles are then resampled according to these updated (normalized) weights. Note that we used an MCMC kernel with an invariant distribution matching the target posterior as the forward kernel. SMC samplers also require the specification of backward kernels. We use the corresponding time-reversal kernel as the backward kernel as outlined in Section 3.3.2.3 in \citet{del2006sequential}.


Before providing the details of the gradient-based kernel, we first argue how we can easily derive an unbiased estimate of the prequential scoring rule. As discussed in Section \ref{sec:sr_post}, the prequential SR $S({P}_{\theta_{st}}(.|s_{t}, a_{t}), s_{t+1})$ can also be expressed as an expectation over samples from $P_{\theta_{st}}$, conditioned on $(s_{t}, a_{t})$ for all $t = 1, 2, \ldots T$. Whenever $S$ is differentiable with respect to $\theta$, an unbiased estimator of the gradient of the total prequential SR $\PS_T(\theta)$ can be obtained using random samples from an auxiliary distribution such as a Gaussian or uniform distribution that is independent of $\theta$ using equation (\ref{mc-esr}). Next, we describe the adjusted stochastic gradient Riemannian Langevin dynamic (adSGRLD) kernel used as the forward kernel for SMC. 

\paragraph{Adjusted stochastic gradient Riemannian Langevin dynamic}
Suppose we wish to sample from a distribution with pdf $p(\theta) = (1/Z) \exp(-U(\theta))$, $\theta \in \mathbb{R}^p$;  where $Z$ is a normalizing constant and $U(\theta)$, the negative log-likelihood equivalent, is known as the potential energy. Standard SGLD is based on the Overdamped Langevin Diffusion, represented by the following stochastic differential equation which has a stationary distribution $p\left(\theta\right)$:
$$
d\theta(u) = -\frac{1}{2} \nabla_{\theta} U(\theta(u))du + dB_u,
$$
where $B_u$ is a Brownian motion. To sample from the target distribution using the above SDE, we often use numerical approximation schemes like the Euler-Maruyama discretization, which leads to the following update rule:
\begin{equation}\label{eqn-sgld}
    \theta_{u+1} \leftarrow \theta_u - \frac{\epsilon}{2} \widehat{\nabla_{\theta}}U(\theta_u) + \sqrt{\epsilon_u} W,
\end{equation}
where $W$ is a d-dimensional standard normal random vector; $\widehat{\nabla_{\theta}}U(\theta_u)$ (often used in practice) is an unbiased estimator of the gradient of $U(\theta_u)$ and $\{\epsilon_u\}$ is a sequence of discretization step sizes satisfying the conditions, $\sum_{u=1}^{\infty} \epsilon_u = \infty$ and $\sum_{u=1}^{\infty} \epsilon_u^2 < \infty$.

Although \citet{welling2011bayesian} have shown that SGLD yields samples from the target posterior when using a sequence $\epsilon_u$ converging to $0$, in practice, $\epsilon_u$ seldom converges to $0$, leading to bias due to Euler-Maruyama discretization. Hence, to ensure random sampling with minimal bias even when using a noisy estimate of the gradient, adaptive Langevin dynamics has been proposed in \citet{jones2011adaptive} which was later adapted for Bayesian inference \citep{ding2014bayesian} and likelihood-free inference \citep{pacchiardi2024generalized}. The algorithm, referred to as adaptive stochastic gradient Langevin dynamics (adSGLD), runs on an augmented space $(\theta, \kappa, \eta)$, where $\theta$ represents the parameter of interest, $\kappa \in \mathbb{R}^p$ represents the momentum and $\eta$ represents an adaptive thermostat controlling the mean kinetic energy $\frac{1}{p} \mathbb{E}(\kappa^T\kappa)$.

Choosing an appropriate sequence of step sizes is crucial for effectively exploring the parameter space, especially in high dimensions. For example, if different components of $\theta$ have values on different scales or the components are highly correlated, a poor choice of step sizes can result in slow mixing, negatively impacting the performance of the SMC sampler. To address this issue, several preconditioning schemes \citep{girolami2011riemann} have been proposed in the literature. For instance, the Riemann manifold Metropolis-adjusted Langevin algorithm uses a positive definite matrix $G(\theta)$ to adaptively precondition the gradient. Then the SDEs are given by,

\begin{eqnarray*}
d \boldsymbol{\theta}&=& G(\theta)\boldsymbol{\kappa} d u,\\
d \boldsymbol{\kappa}&=&\left(-G(\boldsymbol{\theta}) \nabla_{\boldsymbol{\theta}} U(\boldsymbol{\theta})
-\boldsymbol{\eta}\boldsymbol{\kappa}
+ \nabla_{\boldsymbol{\theta}} G(\boldsymbol{\theta})
+ G(\boldsymbol{\theta})(\boldsymbol{\eta} - G(\boldsymbol{\theta})) \nabla_{\boldsymbol{\theta}}G(\boldsymbol{\theta})\right)d u +\sqrt{2} G(\boldsymbol{\theta})^{\frac{1}{2}} d B_u,\\
d \boldsymbol{\eta} &=& \left(\frac{1}{p}\boldsymbol{\kappa}^T\boldsymbol{\kappa} - 1\right)d u
\end{eqnarray*}
where $G(\theta)$\footnote{We used the preconditioning matrix inspired by Adam optimizers as suggested in \cite{chen2016bridging}.} is our preconditioning matrix, and 
$\nabla_{\boldsymbol{\theta}} G(\boldsymbol{\theta})$ is a vector with $i$-th element being $\sum_j \nabla_{\boldsymbol{\theta}_j} G_{ij}(\boldsymbol{\theta})$. $G(\boldsymbol{\theta})$ encodes geometric information of the potential energy $U(\boldsymbol{\theta})$, called Riemannian metric \citep{girolami2011riemann}, which are commonly defined by the Fisher information matrix. 

If we assume $G(\boldsymbol{\theta})$ does not depend on $\boldsymbol{\theta}$, we notice that the third and fourth terms in the momentum update term vanishes and we end up with an SDE of the form
\begin{eqnarray}
\label{eq:rie_adj_sgld}
d \boldsymbol{\theta}&=& G(\theta)\boldsymbol{\kappa} d u,\nonumber\\
d \boldsymbol{\kappa}&=&\left(-G(\boldsymbol{\theta}) \nabla_{\boldsymbol{\theta}} U(\boldsymbol{\theta})
-\boldsymbol{\eta}\boldsymbol{\kappa}\right)d u+\sqrt{2} G(\boldsymbol{\theta})^{\frac{1}{2}} d B_u,\nonumber\\
d \boldsymbol{\eta} &=& \left(\frac{1}{p}\boldsymbol{\kappa}^T\boldsymbol{\kappa} - 1\right)d u
\end{eqnarray}
Using an Euler scheme with step size $\epsilon$ in the set of Equations \ref{eq:rie_adj_sgld}, we obtain the following sampling Algorithm \ref{alg2} where in place of $\nabla_{{\theta}} U({\theta})$, we use its unbiased estimate $\widehat{\nabla_{\theta}}U(\theta_u)$ (similar to equation (\ref{eqn-sgld})).

\begin{algorithm}
\caption{Adjusted stochastic gradient Riemannian Langevin dynamic}
\label{alg2}
\begin{algorithmic}[htbp!]
\Require Parameters $\epsilon, a$
\State Initialize $\boldsymbol{\theta}_{(0)} \in \mathbb{R}^p, \mathbf{\kappa}_{(0)} \sim \mathcal{N}(0, G(\boldsymbol{\theta}_{(0)}\epsilon\mathbf{I})$, and $\boldsymbol{\eta}_{(0)} = a$
\For{$t = 1, 2, \ldots$}
    \State Evaluate $\widehat{\nabla_{\theta}}{U}(\boldsymbol{\theta}_{(t-1)})$ 
    \State $\mathbf{\kappa}_{(t)} = \mathbf{\kappa}_{(t-1)} - (\boldsymbol{\eta}_{(t-1)} \mathbf{\kappa}_{(t-1)} + \widehat{\nabla_{\theta}}{U}(\boldsymbol{\theta}_{(t-1)})G(\boldsymbol{\theta}_{(t-1)}))\epsilon + \mathcal{N}(0, 2a \epsilon G(\boldsymbol{\theta}_{(t-1)}))$
    \State $\boldsymbol{\theta}_{(t)} = \boldsymbol{\theta}_{(t-1)} + \mathbf{\kappa}_{(t)}G(\boldsymbol{\theta}_{(t-1)})\epsilon$
    \State $\boldsymbol{\eta}_{(t)} = \boldsymbol{\eta}_{(t-1)} + (\frac{1}{p} \mathbf{\kappa}_{(t)}^T \mathbf{\kappa}_{(t)} - 1)\epsilon$
\EndFor
\end{algorithmic}
\end{algorithm}


When the target distribution is the scoring rule posterior, $U(\theta) = S(P_{\theta}, \mathbf{y})$, where $P_{\theta}$ is a simulator model proposed for the observations $\mathbf{y}$. SRs such as the energy score can be expressed as $S(P_{\theta}, \mathbf{y}) = \mathbb{E}_{X, X' \sim P_{\theta}} g(X, X', \mathbf{y})$. Moreover, a simulation from the model $P_{\theta}$ can also be represented as $x = h_{\theta}(z)$, with $z \sim Z$, where the distribution $Z$ is independent of $\theta$. \cite{pacchiardi2024generalized} showed that, if both $g$ and $h_{\theta}$ are differentiable, then an interchange of expectation and gradient step produces an unbiased estimate of the gradient $\widehat{\nabla_{\theta}}U(\theta_u)$ using some random draws of $z_i \sim Z$ for $i = 1, 2, \ldots m$. This property is particularly useful in high-dimensional parameter spaces, where efficient exploration requires gradient information.

$$
$$


\section{ETS with generalized posterior}
\label{sec:etsq}

In Section \ref{sec:gbrl}, we have introduced a simulator model-based framework designed to obtain posterior samples when the model likelihood is unknown. The parallelizability of SMC samplers makes this algorithm scalable for higher dimensions and computationally efficient. Following this, we now present a result concerning the convergence of approximate policy iteration methods when integrated with ETS.

Evaluating a policy's performance typically involves measuring regret, which requires calculating the value functions for both the optimal policy and the learned policy. This, in turn, depends on knowing the exact state transition probabilities and reward distributions of the true environment to compute the expectation. However, for many complex tasks, these exact distributions defining the environment’s dynamics are intractable, making it necessary to approximate the value functions from observed interaction data.

Therefore, we provide a convergence result for the ETS-based policy in terms of the difference between the action-value functions of the optimal policy and the learned policy. Additionally, we prove this result for a well-specified case where the expected scoring rule minimizer aligns with the true model parameters.

\begin{theorem}\label{th:regret}

    Let $\mu_1, \mu_2, \ldots $ be the sequence of policies generated through ETS with an approximate policy iteration algorithm after $k$ episodes of interaction with the environment. Also, let $Q^{\bm{\theta^{(k)}}}_{\mu_{j}}$ denote the estimated $Q$ function for the policy $\mu_j$ with $\mu_{j+1}(s) = \arg \max_{a}Q^{\bm{\theta^{(k)}}}_{\mu_{j}} (s,a)$ for all $j = 1,2, \ldots$ . Assuming  $Q_{\mu}^{\theta}$ to be Lipschitz continuous for all possible policies $\mu$ and $\theta \in \Theta$, for the case of a well-specified model we have

$$||Q^* - Q^{\bm{\theta^{(k)}}}_{\mu_{j+1}}||_{\infty}  \leq  \gamma^j ||Q^* - Q^{\bm{\theta^{(k)}}}_{\mu_{1}}||_{\infty} + \sum_{l = 1}^{j}\gamma^{j-l +1} \zeta_l (k, n), $$
    where $\gamma \in [0, 1]$ is a discounting factor used to define the value functions. 
\end{theorem}

A detailed proof of the above theorem can be found in Appendix B. The theorem suggests that, regardless of the number of episodes observed, the sequence of policies obtained from a policy iteration method integrated with ETS progressively converges toward optimal behavior as the iterations increase.  The second term $\zeta_l(k, n)$ involves the samples drawn from the posterior of the model parameters obtained after observing up to $k$ episodes of interaction with the environment. As the episode count $k$ and the number of posterior samples $n$ increase, this term shrinks to $0$ for any $l=1, 2, \ldots$ according to Theorem \ref{th-bvm} as the posterior distribution of the model parameters concentrates around the expected prequential score minimizer.

\section{Simulation studies}
\label{sec:results}

In this section, we demonstrate the application of several model-free policy learning algorithms integrated with ETS, comparing the performance of the ETS-integrated approach with that of the classical model-free method. We begin by presenting results for a finite action space problem, showing both well-specified and misspecified model cases. We then extend the analysis to a problem with continuous action space, where we focus on the misspecified model scenario.

\subsection{Finite action MDP}
\paragraph{Well-specified models}
To demonstrate the ETS algorithm, we use the `inverted pendulum' task from \citet{dimitrakakis2013abc}. Here the agent targets to keep the pendulum upright as long as possible by switching actions. The state of the environment is defined as (the angle, and angular velocity) of the pendulum. There are three possible actions from each state. The actions are the force (in Newtons) applied in a certain direction, the action space is $(+50, 0, -50)$.  The physical equation for the system has 6 parameters: the pendulum mass, the cart mass, the pendulum length, the gravity, the amount of uniform noise, and the simulation time interval. The true value of the parameters are $\theta_0 = (2.0, 8.0, 0.5, 9.8, 10, 0.01)$ which is unknown, however, the simulator model ($P_{\theta}$) for the dynamics of the environment is known to us. In this environment, the agent gets a $+1 $ reward for every balancing step.

In our comparison, we evaluated the online performance of both the model-free and Bayesian model-based approaches in the pendulum domain. In the model-free approach, we updated our policy after each episode of interaction with the environment. The policy updates ceased once we reached the maximum reward, which is known to be $1000$ in our setup because an episode is defined by a maximum of 1000 steps or until the pendulum falls for the first time.

\begin{figure}
    \centering
    \begin{minipage}{0.55\linewidth}
        \includegraphics[width=\linewidth]{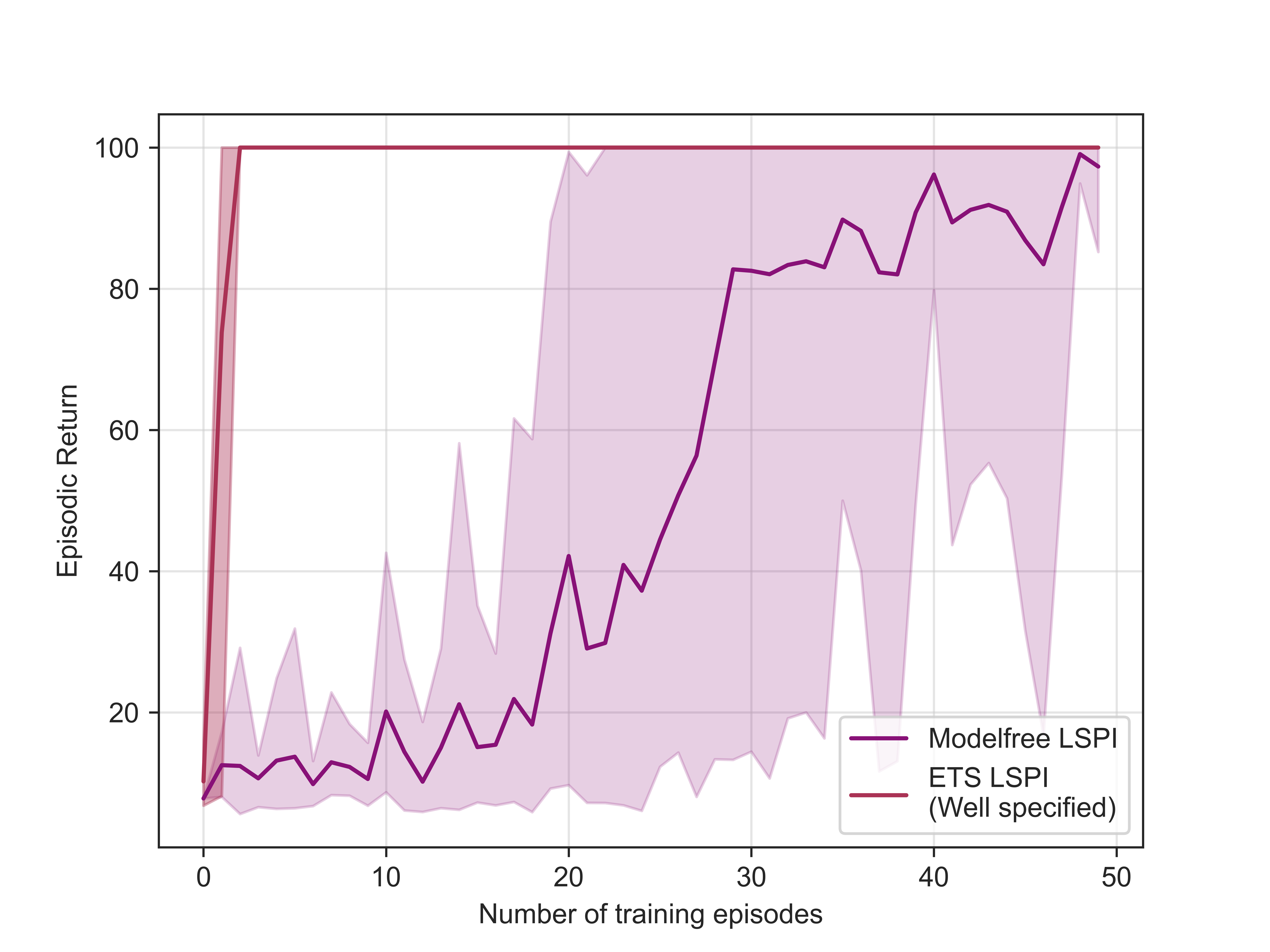}
    \end{minipage}%
    \hfill
    \begin{minipage}{0.4\linewidth}
        \captionof{figure}{Both the policies are trained using the same dataset generated through a random policy. The learned policy is used to interact with the environment for a maximum of 1000 steps or until the pendulum falls for the first time. The same experiment is run independently 10 times and the average discounted return is plotted.}
        \label{pend}
    \end{minipage}
\end{figure}
    
For policy learning, we adopted the least-squares policy iteration (LSPI) method proposed by \cite{lagoudakis2003least}. LSPI is a model-free approximate policy iteration method that leverages the least squares temporal-difference learning algorithm to approximate the $Q$ function. Specifically, we approximated the $Q$ function using a linear combination of a $4 \times 4$ grid of Gaussian radial basis functions in LSPI, with a learning rate of $\gamma = 0.99$.

Additionally, for the Bayesian approach, we assumed a uniform prior over the interval $[0.5 \theta_0, 5.0 \theta_0]$ for the model parameters. We obtained samples from the posterior distribution after each episode using SMC with the gradient-based Markov kernel \citep{jones2011adaptive} (refer to Table 1 in Appendix C for details of the hyperparameters). To define the intermediate target distributions between consecutive episodes, we chose the sequence of temperatures  $\{\alpha_{k,l}\}_{l=1}^L$ at the $k$-th episode, such that the effective sample size (ESS) declines uniformly throughout the SMC iterations. We used the bisection method to find the temperature, $\alpha_{k,l}$ such that $\text{ESS}_l = c_0 \times \text{ESS}_{l-1}$, the parameter $c_0$ for each experiment is mentioned in Appendix C.1. Once the ESS drops below half the original sample size, the particles are resampled. Here we used the zeroth order gradient (discussed in Appendix C.2) to compute an estimate of the gradient of the prequential loss. We used LSPI integrated with ETS to learn an optimal policy, as described in Section \ref{sec:exp_TS} where we updated the policy for a maximum of $J = 30$ iterations in between each episode.


Figure \ref{pend} illustrates that both the model-free and Bayesian model-based methods exhibit improved policy learning with increased training data. However, the Bayesian approach achieves optimal rewards much sooner compared to model-free training. This expedited convergence in Bayesian methods can be attributed to the fact that once the posterior distribution of the model parameters converges, simulated interactions closely resemble true interactions. With this concentrated posterior distribution, longer trajectories of simulated data can be generated, facilitating a more accurate estimation of the value function. Consequently, when using these longer chains of simulated data to estimate the value function and perform LSPI to find an optimal policy, convergence occurs more rapidly due to the utilization of a more stable estimate of the $Q$-function.

\paragraph{Misspecified models}\label{par:ms pend}
 To demonstrate the application of ETS for the case where the simulator model is not known, we revisit the `inverted pendulum' experiment with the previously mentioned parameters. Conditional GANs \citep{charlesworth2020plangan,zhao2021model} have been widely adopted in the literature to model the dynamics. We have used a generative neural network, skipping the adversarial training, to model the difference between the next state and the current state conditioned on the current state and action as suggested by \cite{nagabandi2018neural} and \cite{deisenroth2013gaussian}. Further, for the states $s$ that represents the angle, we have considered feeding in $(\sin(s), \cos(s))$ as inputs to the model. The model can be written as,
$$
f_{\theta}(s_t, a_t, z) = s_{t+1} - s_t
$$
 where $z$ is a Gaussian noise and $f_{\theta}$ is defined by $3$ fully connected layers, $10$ neurons per layer and `swish' activation functions \citep{ramachandran2017searching} following the architecture suggested by \citet{chua2018deep}. 
 
 For parameter inference we have used Sequential Monte Carlo (SMC) (implementation details can be found in Appendix C.1). In high-dimensional parameter spaces, selecting a well-chosen initial sample from an informative prior is crucial for effective Bayesian inference. To do so, we ran the Adam optimizer \citep{kingma2014adam} with a learning rate of $0.001$ for $1000$ steps, saving the final $100$ points. The covariance of these optimized points was computed and used to add Gaussian noise, generating a total of $300$ particles to initialize the SMC.  To define the intermediate target distributions between consecutive episodes, we chose the sequence of temperatures  $\{\alpha_{k,l}\}_{l=1}^L$ at the $k$-th episode, such that the conditional effective sample size (CESS) \citep{zhou2016toward} stays constant throughout the SMC iterations. We used the bisection method to find the temperature, $\alpha_{k,l}$ such that $\text{ESS}_l = c_0 \times N$, where $N$ is the sample size and the parameter $c_0$ for each experiment is mentioned in Appendix C.1. Once the ESS drops below $N/2$, the particles are resampled.

\begin{figure}
    \centering
    \begin{minipage}{0.55\linewidth}
        \includegraphics[width=\linewidth]{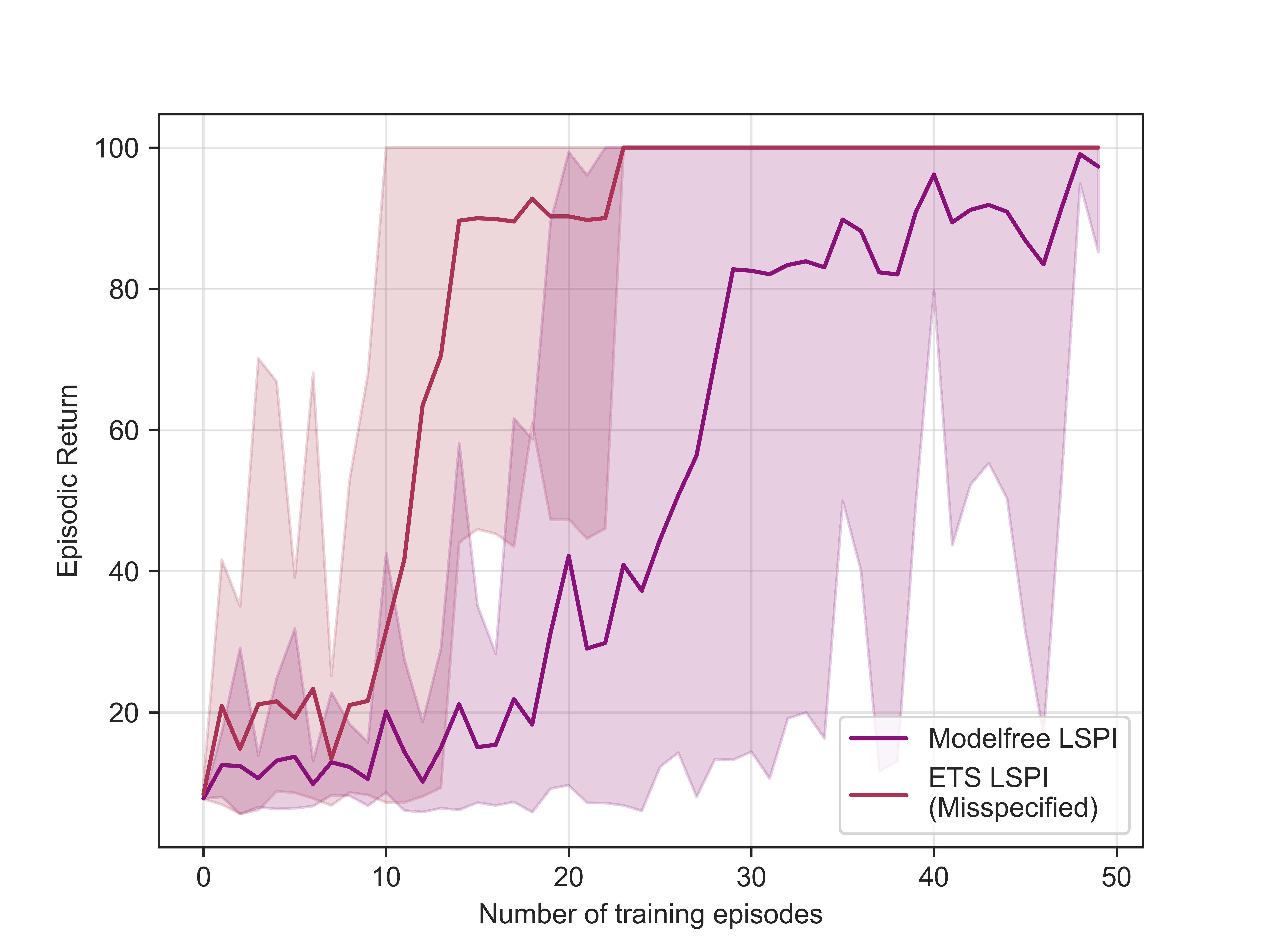}

    \end{minipage}%
    \hfill
    \begin{minipage}{0.4\linewidth}
        \captionof{figure}{Both the policies are trained using the same dataset generated through a random policy. The learned policy is used to interact with the environment for a maximum of 1000 steps or until the pendulum falls for the first time. The same experiment is run independently 10 times and the average discounted return is plotted.}
        \label{pend_nn}
    \end{minipage}
\end{figure}

Figure \ref{pend_nn} demonstrates that LSPI combined with ETS learns the optimal policy significantly faster than its model-free counterpart. Although the true dynamics of the environment were unknown, the generative neural network effectively learned the dynamics, accelerating the policy learning process. Note that the Theorem \ref{th-bvm} on the consistency of the generalized posterior distribution and the Theorem \ref{th:regret} on the bound on the approximation error of the $Q$ function under ETS assume an unimodal posterior for model parameters. While this assumption may not hold when using a generative neural network as a simulator model, we still demonstrate that ETS is more sample-efficient for simpler models.

SMC performance generally benefits from an increased number of particles, which raises computational costs. Hence, we used only $300$ particles for SMC and resampled $100$ particles out of them based on their weights for ETS. This streamlined approach still yielded promising results.

\subsection{Continuous action MDP}

When the action space is continuous, calculating the $Q$-function for all possible state-action pairs is impossible. A deterministic policy that maximizes the $Q$-function over the entire action space cannot be easily found. In such cases, a parameterized probabilistic policy is used, where $\mu_{\alpha}(a|s)$ denotes the probability of taking action $a$ when the environment is in state $s$, and $\alpha \in \mathbb{R}^{d'}$ are the policy parameters.
Usually, the policy distribution is considered as a Gaussian distribution and $\alpha$ would contain the mean and covariance of the distribution. The agent here tries to update the policy parameters according to the rewards collected during real-time interaction with the environment.

In classical policy gradient methods, a performance measure $J(\alpha)$ is computed from real interaction data, and the policy parameters are updated using gradient ascent according to:
 $$
 \alpha_{t+1} \leftarrow \alpha_{t} + \widehat{\nabla} J(\alpha_t),
 $$
 where $\widehat{\nabla} J(\alpha')$ is an estimate of the gradient of $J$ with respect to $\alpha$ evaluated at $\alpha'$. 
 
 In ETS, the performance measure of a policy is estimated from simulated interactions based on each posterior sample of the model parameters. Let $J^{\theta}(\alpha)$ represent the performance measure of the policy $\mu_{\alpha}$ based on interactions simulated from the MDP $\mathbb{M}_{\theta}$. The pooled estimate of $J$ for policy $\mu{\alpha}$, based on posterior samples $\bm{\theta}^{(k)}$, is given by:
\begin{equation}\label{eq:ets-j}
    J^{\bm{\theta}^{(k)}} (\alpha) = \frac{1}{n} \sum_{i=1}^{n} J^{\theta_{ki}}(\alpha) \approx \int_{\Theta} J^{\theta}(\alpha) \pi_k(\theta) d\theta. 
\end{equation}
Then, the policy parameters $\alpha$ are updated as 
\begin{equation}\label{eq:policygrad}
    \alpha_{t+1} \leftarrow \alpha_{t} + \widehat{\nabla} J^{\bm{\theta}^{(k)}}(\alpha_t).
\end{equation}

We perform the policy update based on the simulated trajectories until convergence and use the latest policy to interact with the environment in the next episode. Although we do not present theoretical results for ETS applied to continuous action spaces, our empirical findings align closely with the theoretical results we previously established for discrete action spaces.

\paragraph{Example} To demonstrate the ETS strategy on a problem with continuous action space for the case of a misspecified model, we choose the `Hopper' experiment from the OpenAI Gymnasium package \citep{kwiatkowski2024gymnasiumstandardinterfacereinforcement}. This problem involves moving a two-dimensional, single-legged structure forward by applying force to three joints connecting its four body parts. Actions here consist of torques (ranging from -1 Nm to 1 Nm) applied to each of the three joints, while the environment’s state is defined as a $12$-dimensional real-valued vector of joint angles, angular velocities, positions, and velocities of the body parts. The reward function incentivizes the forward movement of the hopper while penalizing the application of excessive torque, which could destabilize the system.

We have used a similar neural network architecture as described in Section \ref{par:ms pend}, for our simulator model, using three fully connected layers with $20$ nodes each. For sampling from this posterior of the model parameters, we used SMC, as before, but to maintain computational efficiency we used only $45$ particles (this figure was chosen based on the runs with 100 and 500 particles, which did not provide any significant improvements while already being expensive) for this problem. As before, we ran the Adam optimizer \citep{kingma2014adam} with a learning rate of $0.001$ for $1000$ steps, saving the final $15$ points. The covariance of these optimized points was computed and used to add Gaussian noise, generating a total of $45$ particles to initialize the SMC. Similar to the misspecified `inverted pendulum' problem, we defined intermediate target distributions for SMC based on the CESS. 

For policy learning, we applied the `REINFORCE' \citep{williams1992simple} a foundational policy gradient method suitable for continuous action spaces, integrating ETS via equation (\ref{eq:ets-j}). The policy parameters were then updated based on equation (\ref{eq:policygrad}). Full implementation details of the experiment can be found in Sections C.1 and C.2 in the Appendix.
Given that with an increasing amount of data, the posterior distribution can become overly concentrated, we limited posterior updates and sampling to the first $15$ episodes.  At episode $15$, we halted model training and proceeded with the standard REINFORCE updates to refine the policy during future interactions with the environment.

\begin{figure}
    \centering
    \begin{minipage}{0.5\linewidth}
        \includegraphics[width=\linewidth]{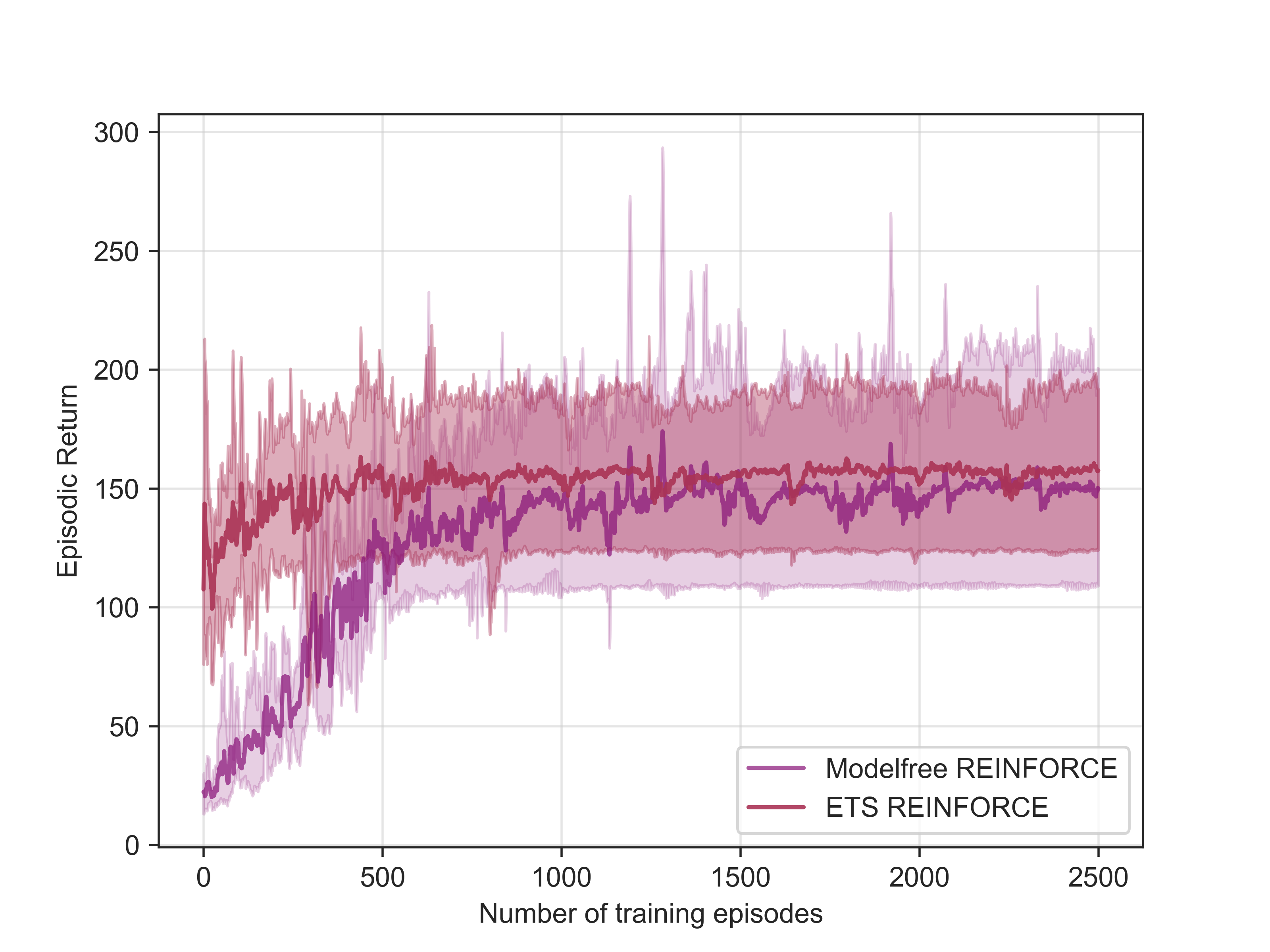}
    \end{minipage}%
    \hfill
    \begin{minipage}{0.4\linewidth}
        \captionof{figure}{For visual clarity we have considered the moving average (lag 10) of episodic returns collected using model-free REINFORCE and ETS-integrated REINFORCE. The average of the returns over five different seeds is plotted for both cases.}
        \label{fig:hopper}
    \end{minipage}
\end{figure}

In Fig. \ref{fig:hopper}, we compare the performance of REINFORCE integrated with ETS versus the classical model-free approach, where we notice that
REINFORCE combined with ETS quickly learns high-rewarding policies, outperforming its model-free counterpart in terms of sample-efficiency. Although model misspecification is present, the universal approximation theorem \citep{hornik1989multilayer, hornik1991approximation} suggests that this error can be minimized by carefully selecting the complexity of the neural networks. Hence the advantage of ETS-integrated policy learning arises from its generative neural network-based model-learning component, which accelerates policy convergence.

\section{Conclusion}
\label{sec:conclusion}

In this work, we have introduced a robust Bayesian framework for model-based reinforcement learning. A fully Bayesian treatment of model parameters traditionally requires a tractable likelihood function of parameters. This is often unavailable when the environment's dynamics are modeled through generative neural networks, but is easier to simulate from. To address this issue, we formulated a generalized posterior for the model parameters using prequential scoring rules based on a Markovian assumption on observed trajectories, enabling generalized Bayesian inference in the absence of a known likelihood. Additionally, we established a Bernstein-von Mises (BvM) type consistency result for the proposed prequential scoring rule posterior in the discrete action setting, leaving the proof for the continuous action space for future work.

For efficient sampling from the generalized posterior, we use SMC samplers. 
We use an adjusted SGLD kernel as the forward kernel in SMC, which handles noisy gradient estimates of the potential. To further improve sampling efficiency, we use gradient-based preconditioning, similar to the Adam optimizer, to better guide particles toward high-probability regions of the parameter space. For policy learning, we extended the classical TS by incorporating full posterior samples for enhanced policy search, introducing the expected Thompson sampling (ETS) approach. Using the BvM result, we derived an error bound to approximate the $Q$-function when a policy iteration method is integrated with ETS in well-specified model settings.

To empirically evaluate the proposed method, we first compared classical model-free LSPI with the ETS-integrated Bayesian version on a simple inverted pendulum-balancing task. In both well-specified and misspecified model settings, the ETS-integrated approach learned the optimal policy more quickly than the model-free baseline. We then applied ETS to the more complex problem of teaching a single-legged hopper to move forward without falling. Even with continuous action space and a misspecified model, the ETS-based approach discovered a high-reward policy significantly faster than the model-free approach. 

In conclusion, we present a robust framework for policy learning that enables rapid identification of high-reward policies in complex tasks. This approach is especially relevant for applications in the design of clinical trials, robotics, and autonomous systems where agents must make quick decisions without the luxury of prolonged learning periods. However, we note that, due to the use of deep generative neural networks as a model for the MDP underlying the environment, the posterior of model parameters can be multimodal. This makes sampling difficult, especially in high dimensions. This may be dealt with using more informative priors (e.g., shrinkage priors) on the parameter space, which we keep as a future direction to explore. 
Another promising avenue could be through imposing a posterior distribution on the parameters of the policy in addition to the model parameters, hence learning the model parameters and optimizing the policy concurrently.

\bibliography{paper.bib}

\appendix

\section{Proofs related to the asymptotic behaviour of generalized posterior}
\label{sec:appendix}

\subsection{Proof of Lemma \ref{lem-limiting_exp-loss}}\label{appedix-lemma1}

\textbf{Lemma \ref{lem-limiting_exp-loss}}Under assumption \ref{assumption1}, there exists a function \( \PS^*(\theta) \) such that as \( T \to \infty \), \( \frac{1}{T}\widetilde{\PS}_T(\theta) \to \PS^*(\theta) \) uniformly with probability one under $\mathbb{M}_0$. 
\begin{proof}
     We can define a statistical divergence between the proposed model $\mathbb{M}_{\theta}$ and the true distribution $\mathbb{M}_0$ in terms of the expected prequential scores upto time $T$ when the associated scoring rule is strictly proper (by assumption \ref{assumption1}). We denote it by
$$D_T(\theta) =\widetilde{\PS}_T(\mathbb{M}_{\theta}, \mathbb{M}_{0}) - \widetilde{\PS}_T(\mathbb{M}_{0}, \mathbb{M}_{0}),$$
where $\widetilde{\PS}_T(\mathbb{M}_{0}, \mathbb{M}_{0})$ is the generalized entropy associated with the model $\mathbb{M}_{0}$. Since the divergence \( D_T(\theta) \) is of order \( T \) \citep{Dawid_2014}, $\frac{1}{T}D_T(\theta)$ has a finite limit and we assume that,
$$
\lim_{T \to \infty} \frac{1}{T}D_T(\theta) = D^{*}(\theta). 
$$

Again from assumption \ref{assumption1} we have $\frac{1}{T}\widetilde{\PS}_T(\mathbb{M}_{0}, \mathbb{M}_{0})$ converging to a constant $c^*$ (say) as $T \to \infty$. Then, 
\begin{align*}
    \lim_{T \to \infty} \frac{1}{T}\widetilde{\PS}_T(\theta) = \lim_{T \to \infty} \frac{1}{T}\widetilde{\PS}_T(\mathbb{M}_{\theta}, \mathbb{M}_{0}) &= \lim_{T \to \infty} \frac{1}{T}\left(D_T(\theta) + \widetilde{\PS}_T(\mathbb{M}_{0}, \mathbb{M}_{0})\right)\\ 
    &= D^{*}(\theta) + c^* = \PS^*(\theta).
\end{align*}
\end{proof}

\subsection{Proof of Lemma \ref{lem-ulln}}
\label{appendix-lem-ulln}

\textbf{Lemma \ref{lem-ulln} }(Uniform law of large numbers) Under assumptions \ref{assumption2} and \ref{assumption3}, a uniform law of large numbers (ULLN) holds, which implies with probability $1$ under $\mathbb{M}_0$,
    $$
    \sup_{\theta \in \Theta}  |\PS_T(\theta) - \widetilde{\PS}_T (\theta)| \to 0 \text{ as } T \to \infty.
    $$

\begin{proof}
We obtain a ULLN from the stationarity, mixing and moment boundedness conditions in assumptions \ref{assumption2} and \ref{assumption3} from a result in \cite{potscher1989uniform} as adapted by \cite{pacchiardi2024probabilistic}. According to the ULLN, for all actions $a \in \mathcal{A}$ with probability $1$ under $\mathbb{M}_0$,  

\begin{equation*}\label{eqn-ulln_action}
    \sup_{\theta \in \Theta} |\PS_T^a(\theta) - \widetilde{\PS}_T^a (\theta)| \to 0,
\end{equation*}
where $\widetilde{\PS}_T^a (\theta) = \mathbb{E}_{H_T^a \sim \mathbb{M}_0} \PS(\mathbb{M}_{\theta}, H_T^a)$ denotes the expected prequential score calculated with respect to the observations where action $a$ was chosen. Then, using the triangle inequality we get with probability $1$ under $\mathbb{M}_0$, 
\begin{align} \label{eqn-ulln}
    \sup_{\theta \in \Theta} |\PS_T(\theta) - \widetilde{\PS}_T (\theta)| &= \sup_{\theta \in \Theta} |\sum_{a \in \mathcal{A}}\PS_T^a(\theta) - \sum_{a \in \mathcal{A}}\widetilde{\PS}_T^a (\theta)|\notag \\
    &\leq \sup_{\theta \in \Theta} \sum_{a \in \mathcal{A}} |\PS_T^a(\theta) - \widetilde{\PS}_T^a (\theta)| \to 0.
\end{align}
\end{proof}

\subsection{Proof of Corollary \ref{cor}}
\label{appendix-cor}
\textbf{Corollary \ref{cor}} Under assumptions \ref{assumption1}-\ref{assumption3}, it follows from Lemma \ref{lem-limiting_exp-loss} and Lemma \ref{lem-ulln} that, 
    $$
    \sup_{\theta \in \Theta}  |\frac{1}{T}\PS_T(\theta) - {\PS}^* (\theta)| \to 0 \text{ as } T \to \infty,
    $$
    with probability $1$ under $\mathbb{M}_0$.

\begin{proof}
    Under assumption \ref{assumption1}, from Lemma \ref{lem-limiting_exp-loss}, for a fixed $\epsilon >0$ there exists a $T_1(\epsilon)$ such that for all $T > T_1(\epsilon)$ with probability $1$ under $\mathbb{M}_0$, 
\begin{align}\label{eqn-sup1}
    &|\frac{1}{T} \widetilde{\PS}_T(\theta) - \PS^*(\theta)| < \epsilon/2, \text{ for all } \theta \in \Theta \notag\\
    &\implies \sup_{\theta \in \Theta} |\frac{1}{T} \widetilde{\PS}_T(\theta) - \PS^*(\theta)| < \epsilon/2.
\end{align}
Similarly, under assumptions \ref{assumption2} and \ref{assumption3}, Lemma \ref{lem-ulln} implies that for the fixed $\epsilon >0$ there exists a $T_2(\epsilon)$ such that for all $T > T_2(\epsilon)$ with probability $1$ under $\mathbb{M}_0$, 
\begin{equation}\label{eqn-sup2}
    \sup_{\theta \in \Theta} | {\PS}_T(\theta) - \widetilde{\PS}(\theta)| < \epsilon T/2.
\end{equation}
On combining equations \ref{eqn-sup1} and \ref{eqn-sup2}, for any $T > \max \{T_1(\epsilon), T_2(\epsilon)\}$ 
\begin{align}\label{eqn-ulln2}
    \sup_{\theta \in \Theta} |\frac{1}{T} {\PS}_T(\theta) - \PS^*(\theta)| &\leq \sup_{\theta \in \Theta} \left\{|\frac{1}{T} {\PS}_T(\theta) - \frac{1}{T}\widetilde{\PS}_T(\theta)| +  |\frac{1}{T} \widetilde{\PS}_T(\theta) - \PS^*(\theta)|\right\} \notag\\
    &< \epsilon/2 + \epsilon/2 = \epsilon,
\end{align}
with probability $1$ under $\mathbb{M}_0$.
Since equation \ref{eqn-ulln2} holds for an arbitrary $\epsilon >0$, it follows that, with probability $1$ under $\mathbb{M}_0$,
$$
\sup_{\theta \in \Theta} |\frac{1}{T} {\PS}_T(\theta) - \PS^*(\theta)| \to 0 \text{ as } T \to \infty.
$$

\end{proof}

\subsection{Proof of Lemma \ref{lem-consistent_estimators_star}}
\label{appendix-lem-consistent_estimators_star}
    \textbf{Lemma \ref{lem-consistent_estimators_star} }(Asymptotic consistency) Under assumptions \ref{assumption1}-\ref{assumption5}, as $T \to \infty$, we have with probability $1$ under $\mathbb{M}_0$,
    $$
    d(\hat{\theta}_T, \theta^*) \to 0.
    $$
We present here a proof of the lemma based on Theorem 5.1 in \cite{skouras1998optimal} as adapted by \cite{pacchiardi2024probabilistic}.

\begin{proof}
From assumption \ref{assumption5}, for a fixed $\epsilon >0$ it is possible to find a $\delta(\epsilon) > 0 $ such that, 
\begin{equation}\label{eqn-delta11}
    \min_{\theta: d(\theta, \theta^*) \geq \epsilon} {\PS}^*(\theta) - {\PS}^*(\theta^*) = \delta(\epsilon),
\end{equation}
with probability $1$ under $\mathbb{M}_0$.

Due to Corollary \ref{cor}, with probability $1$ under $\mathbb{M}_0$, there exists a $T_1(\delta(\epsilon))$ such that for all $T > T_1(\delta(\epsilon))$
\begin{equation*}
    |\frac{1}{T} {\PS}_T(\theta^*) - \PS^*(\theta^*)| < \delta(\epsilon)/2,
\end{equation*}
which implies
\begin{align}\label{eqn-delta1}
    \PS^*(\theta^*) &> \frac{1}{T} {\PS}_T(\theta^*) - \delta(\epsilon)/2 \notag\\
    & \geq \frac{1}{T} {\PS}_T(\hat{\theta}_T) - \delta(\epsilon)/2,
\end{align}
where the second inequality is from the definition of $\hat{\theta}_T$.

On exploiting Corollary \ref{cor} once again, we can define a $T_2(\delta(\epsilon))$ such that for all $T > T_2(\delta(\epsilon))$
\begin{equation}\label{eqn-delta2}
    |\frac{1}{T} {\PS}_T(\hat{\theta}_T) - \PS^*(\hat{\theta}_T)| < \delta(\epsilon)/2,
\end{equation} 
with probability $1$ under $\mathbb{M}_0$.

Then, with probability $1$ under $\mathbb{M}_0$, for all $T > \max\{T_1(\delta(\epsilon)), T_2(\delta(\epsilon))\}$ 
\begin{align}\label{eqn-delta3}
    \PS^*(\hat{\theta}_T) - \PS^*(\theta^*) &= \PS^*(\hat{\theta}_T) - \frac{1}{T} {\PS}_T(\hat{\theta}_T) + \frac{1}{T} {\PS}_T(\hat{\theta}_T) - \PS^*(\theta^*)\notag \\
    & < \delta(\epsilon)/2 + \delta(\epsilon)/2 = \delta(\epsilon)
\end{align}
from equation \ref{eqn-delta1} and equation \ref{eqn-delta2}.

Note that equation \ref{eqn-delta3} ensures that the difference considered in equation \ref{eqn-delta11} is smaller than $\delta(\epsilon)$ when $\theta = \hat{\theta}_T$. However, by equation \ref{eqn-delta11}, the same difference is at least $\delta(\epsilon)$ for all $\theta$ that are outside the $\epsilon$-radius ball around $\theta^*$. This implies that, $\hat{\theta}_T$ must lie inside the $\epsilon$-radius ball, meaning that $d(\hat{\theta}_T, \theta^*) < \epsilon$ with probability $1$ under $\mathbb{M}_0$. Since this is true for any $\epsilon > 0$, it follows that, with probability $1$ under $\mathbb{M}_0$
$$
d(\hat{\theta}_T, \theta^*) \to 0\text{ as } T \to \infty.
$$
   
\end{proof}

\subsection{Sketch proof of BvM theorem}\label{appendix-bvm}
Here we present a sketch proof for the BvM theorem (Theorem \ref{th-bvm}) based on some results from \cite{JMLR:v22:20-469}.

\begin{proof}
From Lemma \ref{lem-consistent_estimators_star} we have with probability one under $\mathbb{M}_0$, as $T \to \infty$, 
$$
d(\hat{\theta}_T, \theta^*) \to 0,
$$
which proves the existence of a sequence of estimators $\{\hat{\theta}_T\}_{T=1}^{\infty}$ such that as $T \to \infty$, with probability one under $\mathbb{M}_0$, $d(\hat{\theta}_T, \theta^*) \to 0$ under the assumptions \ref{assumption1}-\ref{assumption5}.

By Theorem 6 from \cite{JMLR:v22:20-469}, using the above sequence of estimators $\{\hat{\theta}_T\}_{T=1}^{\infty}$ along with the assumptions \ref{assumption7}-\ref{assumption9}, $\PS_T(\theta)$ can be represented as,
\begin{equation}\label{eq-taylorseries}
    \PS_T(\theta) = \PS_T(\hat{\theta}_T) + \frac{1}{2} (\theta - \hat{\theta}_T)'H_T (\theta - \hat{\theta}_T) + r_T(\theta - \hat{\theta}_T)
\end{equation}

where $H_T = \PS_T''(\hat{\theta}_T) \in \mathbb{R}^{p \times p}$ is symmetric and $H_T \to  H^*$. There exists $\epsilon_0, c_0 >0$ such that, for all $T$ sufficiently large, for all $\theta \in B_{\epsilon_0}(\mathbf{0})$, we have $|r_T(\theta)| \leq c_0|\theta|^3. $ 

Finally, on using the Taylor series expansion of the function $\PS_T(\theta)$ in equation \ref{eq-taylorseries}, together with the assumptions \ref{assumption6} and \ref{assumption10}, the Theorem \ref{th-bvm} holds (by Theorem 4 from \cite{JMLR:v22:20-469}). 

\end{proof}

For completeness, we report the full statements of Theorem 4 and Theorem 6 from \cite{JMLR:v22:20-469} below as Theorem \ref{th-miller4} and Theorem \ref{th-miller6}, respectively. 

\begin{theorem}[Theorem 4 from \cite{JMLR:v22:20-469}]\label{th-miller4}

Fix $\theta^* \in \mathbb{R}^p$ and let $\pi: \mathbb{R}^p \rightarrow \mathbb{R}$ be a probability density with respect to Lebesgue measure such that $\pi$ is continuous at $\theta^*$ and $\pi\left(\theta^*\right)>0$. Let $\PS_T: \mathbb{R}^p \rightarrow \mathbb{R}$ for $T \in \mathbb{N}$ and assume:

\begin{enumerate}[label=\textbf{M\arabic*}]
    \setcounter{enumi}{0}
    \item\label{assumption-miller1} $L_T$ can be represented as
$$
\PS_T(\theta) = \PS_T(\hat{\theta}_T) + \frac{1}{2} (\theta - \hat{\theta}_T)'H_T (\theta - \hat{\theta}_T) + r_T(\theta - \hat{\theta}_T)
$$
where $\hat{\theta}_T \in \mathbb{R}^p$ such that $\hat{\theta}_T \rightarrow \theta^*, H_T \in \mathbb{R}^{p \times p}$ symmetric such that $H_T \rightarrow H^*$ for some positive definite $H^*$, and $r_T: \mathbb{R}^p \rightarrow \mathbb{R}$ has the following property: there exist $\varepsilon_0, c_0>0$ such that for all $T$ sufficiently large, for all $\theta \in B_{\varepsilon_0}(\mathbf{0})$, we have $\left|r_T(\theta)\right| \leq c_0|\theta|^3;$

\item For any $\varepsilon>0, \liminf _T \inf _{\theta \in B_{\varepsilon}\left(\hat{\theta}_T\right)^c}\left(\PS_T(\theta)-\PS_T\left(\hat{\theta}_T\right)\right)>0$,
\end{enumerate}
then defining $z_T = \int_{\Theta} \exp(-\PS_T(\theta)) \pi(\theta)d\theta$ and $\pi_{GP}(\theta|y^{T}) = \pi(\theta) \exp(-A_T L_T(\theta))/ z_T$ we have,
    \begin{equation*}
        \int_{B_{\epsilon}(\theta^*)} \pi_{GP}(\theta|y^{T}) d\theta \underset{T \rightarrow \infty}{\longrightarrow} 1 \quad \text{for all } \epsilon >0,
    \end{equation*}
    which means, $\pi_{GP}(\theta|y^{T})$ concentrates at $\theta^*$;
    \begin{equation*}
        z_T \approx \frac{\exp(-A_TL_T(\hat{\theta}_T)) \pi(\theta^*)}{\mid \det H^*\mid^{1/2}} \left(\frac{2\pi}{T}\right)^{d/2}
    \end{equation*}
    as $T \to \infty$ (Laplace approximation), and letting $q_T$ be the density of $\sqrt{T}(\theta - \hat{\theta}_T)$ when $\theta \sim \pi_{GP}(\theta|y^{T})$,
    \begin{equation*}
        \int_{\Theta}\left | q_T(\theta) - \mathcal{N}(\theta \mid \mathbf{0}, {H^*}^{-1}) \right| d\theta \underset{T \rightarrow \infty}{\longrightarrow} 0,
    \end{equation*}
    which implies, $q_T$ converges to $\mathcal{N}(\mathbf{0}, {H^*}^{-1})$ in total variation.
\end{theorem}

\begin{theorem}[Theorem 6 from \cite{JMLR:v22:20-469}]\label{th-miller6}

Let $E \subseteq \mathbb{R}^p$ be open and convex, and let $\theta^* \in E$. Let $L_T: E \rightarrow \mathbb{R}$ have continuous third derivatives, and assume:

\begin{enumerate}[label=\textbf{M\arabic*}]
    \setcounter{enumi}{2}
\item there exist $\hat{\theta}_T\in E$ such that $\hat{\theta}_T \rightarrow \theta^*$ and $L_T^{\prime}(\hat{\theta}_T)=0$ for all $T$ sufficiently large,
\item $L_T^{\prime \prime}\left(\theta^*\right) \rightarrow H^*$ as $T \rightarrow \infty$ for some positive definite $H^*$, and
\item $L_T^{\prime \prime \prime}$ is uniformly bounded;
\end{enumerate}
then, letting $H_T=L_T^{\prime \prime}(\hat{\theta}_T)$, condition \ref{assumption-miller1} is satisfied for all $T$ sufficiently large.
\end{theorem}

\section{Approximation error of the Q function}\label{sec:conv-q}

\begin{proof}
For a fixed episode $k$, we have
\begin{align}\label{eqn:qerror}
    ||Q^* - Q^{\bm{\theta^{(k)}}}_{\mu_{j+1}}||_{\infty} &= ||Q^* - Q^{\theta^*}_{\mu^*_{j+1}} + Q^{\theta^*}_{\mu^*_{j+1}} - Q^{\bm{\theta^{(k)}}}_{\mu_{j+1}}||_{\infty} \notag\\
    & \leq ||Q^* - Q^{\theta^*}_{\mu^*_{j+1}} ||_{\infty}  + || Q^{\theta^*}_{\mu^*_{j+1}} -  Q^{\theta^*}_{\mu_{j+1}} +  Q^{\theta^*}_{\mu_{j+1}} - \frac{1}{n} \sum_{i=1}^n Q^{{\theta}_{ki}}_{\mu_{j+1}}||_{\infty} \notag\\
    & \leq \gamma ||Q^* - Q^{\theta^*}_{\mu^*_{j}}||_{\infty} + ||  Q^{\theta^*}_{\mu^*_{j+1}} - Q^{\theta^*}_{\mu_{j+1}}||_{\infty}+ \frac{1}{n} \sum_{i=1}^n || Q^{\theta^*}_{\mu_{j+1}} -  Q^{{\theta}_{ki}}_{\mu_{j+1}}||_{\infty} \notag\\
    &\quad[\text{using the contraction property of Bellman operator under $\theta^*$}] \notag\\
    & = \gamma ||Q^* - Q^{\theta^*}_{\mu^*_{j}}||_{\infty} + ||  Q^{\theta^*}_{\mu^*_{j+1}} - Q^{\theta^*}_{\mu_{j+1}}||_{\infty} + \frac{1}{n} \sum_{i=1}^n \max_{(s,a) \in \mathcal{S} \times \mathcal{A}}| Q^{\theta^*}_{\mu_{j+1}}(s,a) -  Q^{{\theta}_{ki}}_{\mu_{j+1}}(s,a)| \notag\\
    & \leq \gamma ||Q^* - Q^{\theta^*}_{\mu^*_{j}}||_{\infty} + ||  Q^{\theta^*}_{\mu^*_{j+1}} - Q^{\theta^*}_{\mu_{j+1}}||_{\infty} + \frac{1}{n} \sum_{i=1}^n  ||\theta_{ki} - \theta^*|| \max_{(s,a) \in \mathcal{S} \times \mathcal{A}} K_{\mu_{j+1}}(s,a) \notag\\
    & \quad[\text{assuming $Q_{\mu}^{\theta}$ to be Lipschitz continuous w.r.t $\theta$, $\exists  K_{\mu} < \infty$ for all $\mu$}] \notag\\
    & = \gamma ||Q^* - Q^{\theta^*}_{\mu^*_{j}}||_{\infty} + ||  Q^{\theta^*}_{\mu^*_{j+1}} - Q^{\theta^*}_{\mu_{j+1}}||_{\infty} + \frac{1}{n} \sum_{i=1}^n  ||\theta_{ki} - \theta^*|| K'_{\mu_{j+1}} \notag\\ 
    & \quad [\text{where $K'_{\mu} = \max_{(s,a)} K_{\mu(s,a)}$}] \notag\\
    & = \gamma ||Q^* - Q^{\bm{\theta^{(k)}}}_{\mu_{j}} + Q^{\bm{\theta^{(k)}}}_{\mu_{j}} -Q^{\theta^*}_{\mu_{j}} + Q^{\theta^*}_{\mu_{j}} - Q^{\theta^*}_{\mu^*_{j}}||_{\infty}  + ||  Q^{\theta^*}_{\mu^*_{j+1}} - Q^{\theta^*}_{\mu_{j+1}}||_{\infty} \notag\\
    &\quad +\frac{1}{n} \sum_{i=1}^n  ||\theta_{ki} - \theta^*|| K'_{\mu_{j+1}} \notag\\
    & \leq \gamma ||Q^* - Q^{\bm{\theta^{(k)}}}_{\mu_{j}}||_{\infty} + \gamma||  Q^{\theta^*}_{\mu^*_{j}} - Q^{\theta^*}_{\mu_{j}}||_{\infty} + ||  Q^{\theta^*}_{\mu^*_{j+1}} - Q^{\theta^*}_{\mu_{j+1}}||_{\infty} \notag \\
    & \quad +\gamma|| Q^{\bm{\theta^{(k)}}}_{\mu_{j}} -Q^{\theta^*}_{\mu_{j}}||_{\infty} + \frac{1}{n} \sum_{i=1}^n  ||\theta_{ki} - \theta^*|| K'_{\mu_{j+1}} \notag\\
    & \leq \gamma ||Q^* - Q^{\bm{\theta^{(k)}}}_{\mu_{j}}||_{\infty} + \gamma||  Q^{\theta^*}_{\mu^*_{j}} - Q^{\theta^*}_{\mu_{j}}||_{\infty} + ||  Q^{\theta^*}_{\mu^*_{j+1}} - Q^{\theta^*}_{\mu_{j+1}}||_{\infty} \notag \\
    & \quad + \frac{\gamma}{n} \sum_{i=1}^n  ||\theta_{ki} - \theta^*|| K'_{\mu_{j}} + \frac{1}{n} \sum_{i=1}^n  ||\theta_{ki} - \theta^*|| K'_{\mu_{j+1}} \notag\\
    & = \gamma ||Q^* - Q^{\bm{\theta^{(k)}}}_{\mu_{j}}||_{\infty} + \gamma||  Q^{\theta^*}_{\mu^*_{j}} - Q^{\theta^*}_{\mu_{j}}||_{\infty} + ||  Q^{\theta^*}_{\mu^*_{j+1}} - Q^{\theta^*}_{\mu_{j+1}}||_{\infty}\notag\\
    & \quad + \frac{1}{n} \sum_{i=1}^n  ||\theta_{ki} - \theta^*|| (\gamma K'_{\mu_{j}} + K'_{\mu_{j+1}}) \notag\\
    & =  \gamma ||Q^* - Q^{\bm{\theta^{(k)}}}_{\mu_{j}}||_{\infty} + \gamma \zeta_j (k, n) \notag \\
    & =  \gamma^j ||Q^* - Q^{\bm{\theta^{(k)}}}_{\mu_{1}}||_{\infty} + \sum_{l = 1}^{j}\gamma^{j-l +1} \zeta_l (k, n) \notag \\
\end{align}
where $\zeta_j (k, n) = ||  Q^{\theta^*}_{\mu^*_{j}} - Q^{\theta^*}_{\mu_{j}}||_{\infty} + \frac{1}{\gamma}||  Q^{\theta^*}_{\mu^*_{j+1}} - Q^{\theta^*}_{\mu_{j+1}}||_{\infty} + \frac{1}{n} \sum_{i=1}^n  ||\theta_{ki} - \theta^*|| ( K'_{\mu_{j}} + \frac{1}{\gamma} K'_{\mu_{j+1}})$

Let us define, $Y_k = \theta - \hat{\theta}_{k}$, where $\theta \sim \pi_k$ and $\hat{\theta}_k$ denotes the Scoring rule minimizer obtained after observing $k$th episode. So, according to the Bernstein–von Mises (BvM) theorem, $Y_k = O_p((k\tau)^{-1/2})$ assuming $Y_k$ has a finite expectation. Furthermore, from the consistency results of M-estimators \citep{van2000asymptotic}, under certain regularity conditions,$\hat{\theta}_k$ converges to the expected scoring rule minimizer ($\theta^*$) at the rate of $O_p((k\tau)^{-1/2})$.

For the $n$ samples drawn from the $k$th posterior, if we define $Y_{ki} = ||\theta_{ki} - \hat{\theta_{k}}|| + ||\hat{\theta_{k}} - {\theta^*}||$ then, $Y_{ki} = O_p((k\tau)^{-1/2})$ for all $i = 1, 2, \ldots n$. Hence, $1/ n \sum_{i=1}^n |Y_{ki}| = O_p((n k\tau)^{-1/2})$. Therefore, the last term in the equation (\ref{eqn:qerror}) vanishes with large $k$ and $n$. Thus, compared to the classical TS, for ETS, the last term shrinks $\sqrt{n}$ times faster.

Note that, $ \mu^*_j (s) = \arg \max_{a} Q_{\mu^*_{j-1}}^{\theta^*}(s,a)$ and  $ \mu_j (s) = \arg \max_{a} Q_{\mu_{j-1}}^{\bm{\theta^{(k)}}}(s,a)$. So for large enough $k$ and $n$, from the consistency of posterior mean, we can say that, $Q_{\mu^*_{j}}^{\theta^*} \approx  Q_{\mu_{j}}^{\bm{\theta^{(k)}}}$ for any policy iteration step $j$. Then, the second and third term in equation (\ref{eqn:qerror}) also vanish as $k \rightarrow \infty$ and $n \rightarrow \infty$. 

\end{proof}


\section{Implementation details}
\subsection{Tuning parameters for posterior sampling using SMC}
We provide a list of values of the tuning parameters used for the SMC sampler in Table \ref{tab:smc}.
\label{sec:smc tuning}
\begin{table}
    \centering
    \begin{tabular}{cccc}
        \hline
        Parameter & \makecell{Inverted Pendulum \\ Well specified} & \makecell{Inverted Pendulum\\ Misspecified } & Hopper \\
        \hline
        $\epsilon$(adSGLD step size) & $10^{-2}$ & $10^{-5}$ & $10^{-6}$\\
        $a$ (adSGLD parameter) & $10^{-2}$ & $10^{-4}$ & $10^{-6}$ \\
        $c_0$ (ESS/ CESS multiplier) & $0.9$ & $0.9$ & $0.9$\\
        \makecell{Number of adSGLD moves \\ per iteration of SMC} & $10$ & $20$ & $5$\\
        \makecell{Number of simulations to\\ estimate the SR loss} & $10$ & $10$ & $10$\\
        \hline
    \end{tabular}
    \caption{Tuning parameters for SMC}
    \label{tab:smc}
\end{table}

\subsection{Zeroth order gradient}\label{zo}

Often, simulator models are not differentiable, yet we seek to leverage gradient information for improved sampling. In such cases, a gradient-free optimization technique involving zeroth order (ZO) gradient \citep{liu2020primer} can be used. The multi-point ZO gradient estimate of a function  $f(\theta)$ is defined as,

\begin{equation*}
    \widehat{\nabla_{\theta}} U(\theta) = \frac{1}{\mu b} \sum_{i=1}^{b} [U(\theta + \mu z_i)- f(\theta)]z_i
\end{equation*}
with approximation error $O(\frac{d}{b}) ||\nabla_{\theta} U(\theta)||_2^2 + O(\frac{\mu^2d^3}{b}) + O({\mu^2d})$ \citep{berahas2022theoretical}; where $\{z_i\}_{i=1}^{b}$ denotes $b$ i.i.d. samples drawn from $\mathcal{N}(0, I_{d})$ and $\mu$ is a tuning parameter. The above method produces an unbiased estimate of the gradient of the of the smoothed version of $U$ over a random perturbation $Z \sim \mathcal{N}(0, I_d)$ with smoothing parameter $\mu$, $U_{\mu} (\theta) = \mathbb{E}_{Z}[U(\theta + \mu Z)]$. Intuitively, the final term in the error can be viewed as the bias of the estimate, which tends to grow with the dimension of $\theta$. Meanwhile, the first two terms stem from the variance of the estimate, which can be controlled by augmenting the sample size $b$.
Besides, the gradient estimate gets better as $\mu$ is taken to be small, however, in practice the gradient estimate can be affected by system noise if $\mu$ is too small and so the efficiency of the estimate relies on the careful tuning of the smoothing parameter $\mu$. We set $\mu = 0.0001$ and $b = 30$ for our implementation.



\subsection{Implementation details of REINFORCE}\label{sec:reinforce}

We implement REINFORCE for policy learning by drawing actions from three independent Gaussian distributions. These actions are transformed from $\mathbb{R}$ to $[-1, 1]$ space as the action space for `Hopper' is defined as $[-1, 1]^3$. We train a policy network to predict the mean and standard deviation of the action distributions as a function of the observed state. This network comprises three fully connected layers with 32, 64, and 64 neurons, respectively. The policy parameters are updated using one step of the Adam optimizer (setting the learning rate to be $0.0001$) after observing each episode of interaction data. Also we set the discount factor $\gamma = 0.99$ to define the discounted return. 

When integrating REINFORCE with ETS, the policy network architecture remains unchanged. After each real interaction episode with the environment, we train a simulator model to predict the next state given a state-action pair. Using this simulator, we simulate $500$ interaction episodes, updating the policy at each simulated episode with classical REINFORCE. The updated policy is then used in the next real interaction episode, after which the simulator model is retrained. This process alternates between model learning and policy updates until the $15$th episode.


\end{document}